\documentclass[runningheads]{llncs}

\usepackage{graphicx}
\usepackage{amssymb}
\usepackage[hidelinks]{hyperref}
\usepackage[dvipsnames]{xcolor}
\usepackage{subfigure}
\usepackage{comment}

\begin{document}

\title{Collaborative Interactive Evolution of Art\\in the Latent Space of Deep Generative Models}

\titlerunning{Collaborative Interactive Evolution of Art}

\author{Ole Hall \and Anil Yaman}

\authorrunning{O. Hall \and A. Yaman}

\institute{Vrije Universiteit Amsterdam, 1081 HV Amsterdam, Netherlands\\ \email{ole.moritz.hall@gmail.com, a.yaman@vu.nl}}

\maketitle              

\begin{abstract}
Generative Adversarial Networks (GANs) have shown great success in generating high quality images and are thus used as one of the main approaches to generate art images. However, usually the image generation process involves sampling from the latent space of the learned art representations, allowing little control over the output. In this work, we first employ GANs that are trained to produce creative images using an architecture known as Creative Adversarial Networks (CANs), then, we employ an evolutionary approach to navigate within the latent space of the models to discover images. We use automatic aesthetic and collaborative interactive human evaluation metrics to assess the generated images. In the human interactive evaluation case, we propose a collaborative evaluation based on the assessments of several participants. Furthermore, we also experiment with an intelligent mutation operator that aims to improve the quality of the images through local search based on an aesthetic measure. We evaluate the effectiveness of this approach by comparing the results produced by the automatic and collaborative interactive evolution. The results show that the proposed approach can generate highly attractive art images when the evolution is guided by collaborative human feedback.

\end{abstract}

\section{Introduction}

Artificial intelligence (AI) based approaches to art generation have increased their popularity and received a great deal of attention due to available tools such as Artbreeder~\cite{Artbreeder} and Nvidia Canvas~\cite{park2019semantic}. A debate about the value of such art has developed at the latest since the auction of an AI generated artwork at the renowned auction house Christie’s~\cite{christies}. Frequent critics doubt the creativity and novelty that can be generated by AI, while other voices postulate AI mainly as a potent tool of modern artists~\cite{cetinic2022understanding}. 

Deep generative models such as Generative Adversarial Networks (GANs)~\cite{goodfellow2014generative} play a major role in AI based art generation approaches. The basic idea of a GAN is to generate images that cannot be distinguished from real images. With this setup, however, doubts have been raised if artefacts generated in this way are actually creative or merely attempting to emulate the training material~\cite{elgammal2017can}. On the other hand, the generation process offers a creative space, as the fake images can be generated by sampling from the latent space of learned art representations, providing infinite possibilities of potential images~\cite{cetinic2022understanding}.

Due to this feature, GANs open up many possibilities for human-AI interaction in co-creative processes~\cite{cetinic2022understanding,Grabe2022TowardsAF}. For example, they can inspire designers~\cite{designer1,designer2}, and artists may spend hours generating different random images to find attractive or inspiring outcomes~\cite{BrutforceGAN}. A more promising approach to discover art images is to apply (meta-)heuristic search algorithms such as evolutionary computing (EC)~\cite{eiben2003EC}. Evolutionary computing has been employed to  discover the design space in GANs in different fields of application~(e.g.~\cite{bontrager2018deep,roziere2021inspirational,volz2018mario}), but rarely in the field of art generation~\cite{roziere2020evolgan,Artbreeder}.

In this paper, we employ EC to navigate within the latent space of GANs trained on art images. Firstly, we use a specific GAN architecture, the Creative Adversarial Network (CAN)~\cite{elgammal2017can}, to introduce novel art images. CANs can achieve this because the images generated imitate the real art distribution while deviating from established art styles through a modified loss function that penalises simple categorisation into an existing art style. Then, the vector representations of latent space variables are used to encode individuals in EC. We employ evolutionary operators such as crossover and mutation to generate new individuals from the existing ones. To determine the quality of art images (a.k.a. fitness in EC), we use two metrics: (1) automatic aesthetic, and (2) collaborative interactive human evaluation. 

Automatic aesthetic evaluation is based on Neural Image Assessment (NIMA) \cite{talebi2018nima}, a Convolutional Neural Network trained on an annotated dataset for aesthetic visual analysis (AVA)~\cite{murray2012ava}. We also used the automatic aesthetic evaluation metric as an intelligent mutation operator that aims to perform a local search~\cite{roziere2020evolgan} to improve the quality of the images and accelerate the evolutionary process similar to the approaches used in memetic algorithms~\cite{eiben2003EC}.

Collaborative interactive human evaluation is based on the idea of interactive evolutionary computation (IEC)~\cite{takagi2001IEC} where human evaluations replace the fitness evaluation step in EC. The IEC is particularly suitable in cases such as art generation where the measure of the quality is subjective~\cite{romero2008art}. However, in contrast to the classic IEC setup, here, we consider collaborative evaluations from several participants to account for their subjectivity.

To prevent the generation of very similar images especially within the same generation, we introduce diversity preservation mechanisms. If two similar images are detected, one is replaced by a new randomly generated image. This also intends to accelerate the exploration of latent space, as well as to avoid user fatigue from evaluating many similar images~\cite{eiben2003EC}.

Our results show that this methodology is generally suitable for exploring the latent space of possible art images in a GAN, and it is able to create increasingly attractive images. It was shown that both the automatic as well as the collaborative evolution achieved an increasing fitness over the evolutionary process. In addition, we tested whether the local search leads to improvements beyond a random level in the eyes of human participants, but this was not confirmed. Finally, we investigated whether the results obtained from both the automatic aesthetic and collaborative interactive evolution are in fact perceived as more attractive than randomly generated images. The findings indicate that human guidance is crucial for the evolution of art in order to achieve images that are perceived as more attractive.

\section{Related Work}

\subsection{Generative Adversarial Networks} \label{GAN}

Introduced by Goodfellow et al.~\cite{goodfellow2014generative}, GANs aim to generate artefacts that are indistinguishable from real images. For this purpose, two models are in a competing relationship, allowing for an unsupervised learning approach: the generator is trained to generate fake images that are indistinguishable from real ones, while the discriminator is trained to make the distinction by comparing the generated images with real ones. The training is therefore comparable to a min-max game for two networks. Both the generator and discriminator are designed as deep neural networks. Through mutual feedback, they constantly improve each other, eventually leading the generator to generate artefacts that are difficult to distinguish from real ones. These artefacts can be created from underlying latent variables by random sampling.

Despite impressive visual results, training GANs is considered difficult and partly unstable, and they were initially only able to generate low resolutions~\cite{karras2018pgan}. Therefore, many extensions and variants have been developed since then (\cite{gui2023ganoverview}, for an overview). A structural improvement for working with images was provided by Deep Convolutional Generative Adversarial Networks (DCGANs)~\cite{radford2015unsupervised}, which form the basis of most of today's variants~\cite{gui2023ganoverview}. 

Another interesting extension is the CAN~\cite{elgammal2017can}, which is building up on the DCGAN. Elgammal et al. argue that the classical architecture is merely emulating the training material, but that the reference to and influence by other artists' works is natural. They propose a new architecture, referring to a psychology-based theory from Martindale~\cite{Martindale1990Arousal} that says that creative processes always try to evoke arousal, which can for example be achieved through stylistic breaks. At the same time, however, creative work does not want to let this arousal become too great in order to avoid negative reactions. In their architecture, the goal is to generate creative art by finding a balance between mimicking the real art distribution, while deviating from established styles. This should increase the novelty and creativity of the generated images, which was also supported by different experiments with human participants~\cite{elgammal2017can}. With regard to a collaborative interactive evolution of art images, this could be advantageous in that the participants are less likely to be guided in their evaluations by well-known representatives of different art styles. Therefore, the architecture of the CAN is used for the generative model in this work.

\subsection{Latent Space Exploration}

While theoretically every latent vector in a well-trained GAN leads to an output considered by the discriminator as an element of the target group, there is still a great variation between them. This diversity of possible outputs motivates the interest in exploring the latent space and opens up a creative space for human-AI interaction in co-creative processes. In their framework, Grabe et al.~\cite{Grabe2022TowardsAF} make a distinction between four different interaction patterns in human-AI interaction with GANs such as curation, exploration, conditioning, and evolution.

By curation, they mean the most straightforward method, according to which control over the output is achieved through the selection of training data and the subsequent manual selection of generated artefacts. However, brute-forcing new artefacts and manual cherry-picking provides only a low level of control and allows only for random search~\cite{roziere2021inspirational}, even if it is actually used this way in the everyday life of some artists~\cite{BrutforceGAN}. At the same time, the model choice can also be understood as part of curation, for example the choice of the CAN architecture. 

Although all four types could be subsumed under the umbrella term exploration, Grabe et al. use it to describe the iterative adaptation of generated artefacts, for example by moving a slider. One possibility is the interpolation between different images, for example CREA.blender~\cite{crea} allows the exploration of space between different images. Another possibility is the conscious adaptation along either the latent vectors directly or semantic attributes represented in them. 
Examples of this are found in the alteration of facial features~\cite{shen2020face,zaltron2020face} as well as in adding, deleting, or altering aspects in images~\cite{bau2,bau1}.

Conditioning refers to methods that allow people to determine desired aspects in advance, often referred to as conditional GAN. Examples are drawings or contour images that predefine the later output, which was used in the domain of fashion design~\cite{xin2021object} and in the creation of landscape paintings~\cite{park2019semantic}. In addition, instructions can also be given in text form~\cite{reed2016,designer2}.

\subsubsection{Latent Variable Evolution}

Another form of control is offered by evolutionary computing. Grabe et al.~\cite{Grabe2022TowardsAF} refer under evolution only to explicitly interactive setups since in IEC, human feedback directly guides the process of computer-driven exploration. However, already the definition of a target of the evolutionary process as well as the specification of an automatically evaluating fitness function provide a certain degree of control. 

Evolutionary computing have already been used several times in combination with GANs, both with automatic and interactive fitness evaluation metrics. The terms Latent Variable Evolution~\cite{bontrager2018deepmasterprints,schrum2020interactive,volz2018mario} and, in an interactive setup, Deep Interactive Evolution~\cite{bontrager2018deep} were coined. The first steps in this direction were taken by Bontrager et al.~\cite{bontrager2018deepmasterprints}, who applied an evolutionary search to find a latent vector in a GAN trained on real fingerprints that matches as many subjects as possible. Bontrager et al.~\cite{bontrager2018deep} were also the first to propose an interactive setup for developing faces and shoes towards a target image, thereby demonstrating that GANs can work as a compact and robust genotype-to-phenotype mapping. Again in the field of fashion design, an interactive setup using a conditional GAN with contour images was employed~\cite{xin2021object}. Zaltron et al.~\cite{zaltron2020face} chose an interactive setup with the additional possibility of fine-tuning the faces obtained during the process with the help of sliders. Also in~\cite{roziere2021inspirational} images were developed either by human feedback or by similarity to a target image. It was further shown that genetic algorithms are able to generate diverse sets of latent variables~\cite{fernandes2020latent}, which was also used to augment sparse training datasets~\cite{fernandes2020improv}. In addition, game levels were developed both automatically~\cite{volz2018mario} and interactively~\cite{schrum2020interactive}. In a slightly different approach, Roziere et al.~\cite{roziere2020evolgan} used a (1+1) evolution strategy to find local optima in the neighbourhood of generated artefacts including artworks using automatic image quality evaluation metrics.

\subsection{Evolutionary Art} \label{EvoART}

\subsubsection{Interactive Fitness Evaluation}

Evolutionary art can look back on a long history (\cite{romero2008art}, for an overview). From the beginning, the subjectivity of art has been a major challenge, since the definition of an objective fitness function is not trivial~\cite{machado2008experiments,mccormack2008facing}. For this reason, in various creative domains such as games, music, and image generation often an IEC approach has been chosen~\cite{takagi2001IEC}. One problem with IEC, however, is that humans can only evaluate a relatively small number of candidates in a reasonable amount of time and get tired after only a few generations, which is known as user fatigue. Therefore, numerous attempts have already been made to reduce it~\cite{takagi2001IEC}. 

One solution to these problems can be provided by crowdsourcing based interactive evolution. In image generation, this approach has been applied by Picbreeder~\cite{secretan2008picbreeder}, where people can further evolve the evolved images of others online. This approach is also utilised in Artbreeder~\cite{Artbreeder}, a creative platform for the creation of GAN-based images, which was inspired by Picbreeder.

The IEC approach proposed in this work involves a collaborative approach differing from those already described in that the fitness is formed based on the ratings of several people. Even though only a small number of the same participants evaluated the images in this study, this approach seems well suited in such a subjective domain. 

In general, it opens up interesting possibilities for taking different scores into account and measure the subjectivity. Expanding this approach to different users at different times could also help overcome the problem of user fatigue. An example of this is the Electric Sheep Project~\cite{sheep}, ongoing since 1999, in which the fitness is determined by many different users around the world. 

\subsubsection{Automatic Fitness Evaluation}

Given the problems in IEC, there has been plenty of research on automatic aesthetic fitness evaluation metrics~\cite{machado2008experiments,mccormack2008facing}. Using deep learning methods, great progress has been made for assessing images~\cite{talebi2018nima}. The availability of large annotated datasets for aesthetic visual analysis, such as AVA~\cite{murray2012ava}, which contains over 250,000 images rated by hobby photographers, have also contributed to these advances. 

Even though most datasets consist of photographs, it has already been shown that automatic evaluation metrics can improve artworks for human viewers, at least by using a technical image quality assessment~\cite{roziere2020evolgan}. Despite their less successful results in this way, we chose an aesthetic visual metric, since we assume that art does not only function through its technical quality. The aesthetic quality of the images is determined using Neural Image Assessment (NIMA)~\cite{talebi2018nima}, which is based on an InceptionResNet-v2~\cite{szegedy2016rethinking} image classifier architecture and is trained on AVA~\cite{murray2012ava}.

\section{Methods}

\subsection{Creative Adversarial Network}

In this work, the generator part of a GAN is used as genotype-to-phenotype mapping. As explained in \ref{GAN}, the CAN~\cite{elgammal2017can} architecture is chosen for this. It aims to generate creative art that mimic the real art distribution, but at the same time deviate from established art styles. To achieve this, in addition to the classification of true and fake images, the discriminator has the further objective of assigning the true images to a certain art style. The generator, on the other hand, still has the goal of ensuring that the generated images are not recognised as fake, but in addition the discriminator should find it as difficult as possible to classify them into a particular art style. Further information and a block diagram illustrating this setup can be found in the original work~\cite{elgammal2017can}.

\subsubsection{Technical Details}

The design essentially follows the CAN~\cite{elgammal2017can} architecture, which in turn is based on the DCGAN~\cite{radford2015unsupervised} architecture. This consists of a series of strided convolutions for the discriminator and fractional strided convolutions for the generator. Each convolution is followed by a batch normalisation, except in the generator output layer and in the discriminator input layer. The activation function used in the discriminator is Leaky ReLu and in the generator ReLu, only in the generator output Tanh. A special feature of the CAN architecture is that the last strided convolution in the discriminator is followed by two heads. The first determines the probability of coming from the real image distribution using a fully connected layer. The second determines the probability of classification into the different art styles by means of three fully connected layers. 

Since we showed the images on screens in 16:9 format, the original square format was converted accordingly. For this purpose, in the second, third, and fourth fractional strided convolution in the generator, the kernel sizes are adjusted from (4,4) to (2,4). The latent vectors $z$ that are put under evolutionary control after training are of length 100 and drawn from a standard normal distribution.
The exact architecture is thus:
\\
\\
Generator:\\
$z\in\mathbb{R}^{100} \rightarrow 4\times4\times1024 \rightarrow 8\times6\times1024 \rightarrow 16\times10\times512 \rightarrow 32\times18\times256 \rightarrow 64\times36\times128 \rightarrow 128\times72\times64 \rightarrow 256\times144\times3$ (final resolution)
\\
\\
\\
Discriminator:\\
$256\times144\times3 \rightarrow 128\times72\times32 \rightarrow 64\times36\times64 \rightarrow 32\times18\times128 \rightarrow 16\times9\times256 \rightarrow 8\times4\times512 \rightarrow 4\times2\times512$

head 1: $4\times2\times512 = 4096 \rightarrow 1$

head 2: $4\times2\times512 = 4096 \rightarrow 1024 \rightarrow 512 \rightarrow K$ (number of art styles)
\\
\\
We used the publicly available WikiArts dataset~\cite{tan2018artgan} as training data. It consists of 81,444 artworks from 27 different art styles. The images were normalised and resized to the appropriate resolution. Since this resolution is too low for display, we upsampled the images generated by a factor of eight for human evaluation, resulting in a resolution of $2048\times1152$. The Laplacian Pyramid Super-Resolution Network (LapSRN)~\cite{LapSRN} was used for this purpose. The framework is available for experimentation\footnote{\textcolor{blue}{\url{https://github.com/OMHall/CollaborativeArt}}}.

\subsection{Collaborative Interactive Evolution}

\begin{figure}[h]
    \centering
    \includegraphics[width=0.85\textwidth]{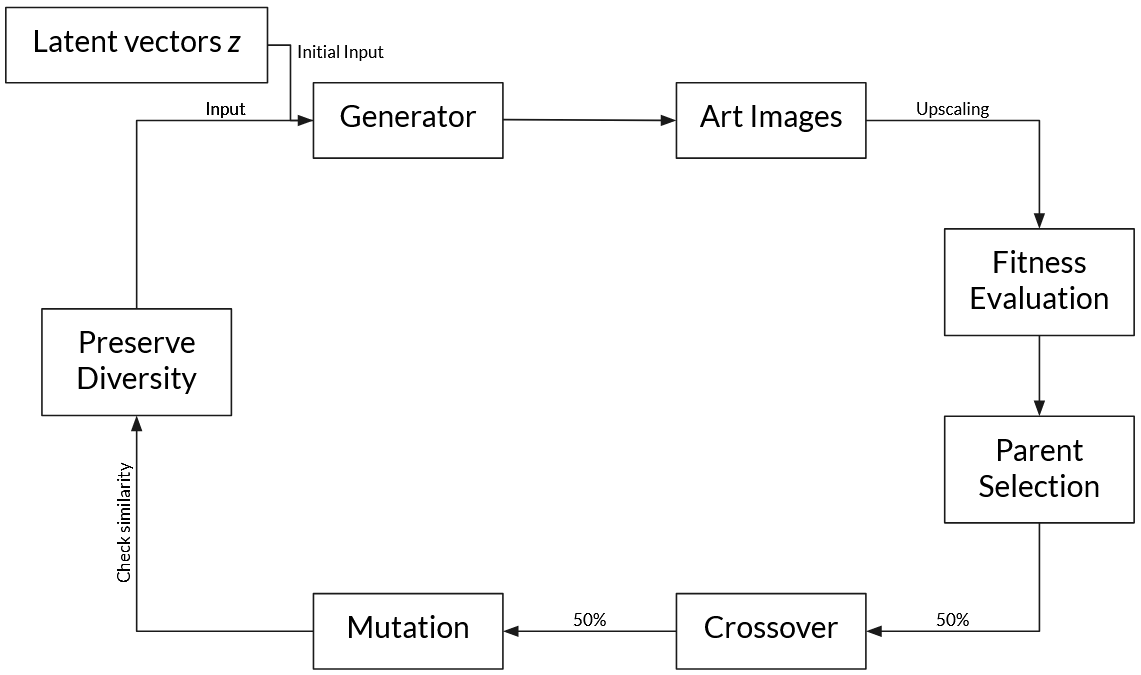}
    \caption{Block diagram of the evolutionary algorithm.}
    \label{fig:EvolutionProcess}
\end{figure}

After training, the generator is able to generate images from every possible latent vector $z$ that follow the distribution of the training images but are difficult to classify into a specific art style. The first generation in the evolution consists of images resulting from randomly generated latent vectors. To not overwhelm the users, IEC classically uses small population sizes and a low number of generations~\cite{bontrager2018deep,mccormack2008facing}. In this work, we choose a population size of 15 and evolve it over 25 generations. In the following, the individual stages of the evolutionary algorithm are outlined. An overview of the algorithm is provided in Figure~\ref{fig:EvolutionProcess}. 

\subsubsection{Fitness Evaluation}
In the collaborative interactive evolution, five participants rated each image independently on a scale of 1 to 10. For this purpose, a questionnaire with the upsampled images was sent to all participants every generation. They were advised to rate the images independently according to how much they like them, rather than comparing them. The average of these ratings then form the fitness of an image. An excerpt of the questionnaire can be found in the supplementary material in Figure~\ref{fig:survey_coll}. 

In the automatic evolution, fitness is evaluated using an automatic aesthetic evaluation by NIMA~\cite{talebi2018nima}, as outlined in~\ref{EvoART}. Starting from the same initial population and using the same algorithm otherwise, this allows for a comparison of the results. 

\subsubsection{Parent Selection}
Due to the collaborative evaluation, no direct selection of parents can be performed as is often the case in IEC~\cite{bontrager2018deep,eiben2003EC,xin2021object}. Instead, stochastic universal sampling (SUS)~\cite{baker1987reducing} is used for parent selection, whereby 15 parents are selected. The next generation is formed from these through crossover and mutation. 

\subsubsection{Crossover}
Crossover is applied to two randomly chosen parents in 50\% of the cases. As in other studies on IEC in combination with GANs, uniform crossover is chosen~\cite{bontrager2018deep,xin2021object}, whereby the probability of exchanging individual genes is set to 25\%. The rationale behind this is that crossover should lead to interesting new images on the one hand, but on the other hand the selected parents should also be recognisable in order to reduce user frustration due to the loss of good solutions~\cite{eiben2003EC}. 

\subsubsection{Mutation}
In order to explore the latent space quickly, the mutation probability is also set to 50\%, as in \cite{bontrager2018deep}. A local search is implemented as mutation, which should both accelerate exploration and further increase the quality of the images in order to keep user fatigue low. This can be understood as intelligent mutation in the sense of a hybrid or memetic evolutionary algorithm~\cite{eiben2003EC}.

The local search is implemented following Roziere et al.~\cite{roziere2020evolgan}, who search for the best possible image in the neighbourhood of a latent vector $z$. For this, a (1~+~1) evolution strategy~\cite{eiben2003EC} is used. In this strategy, there is initially one parent image, from which one offspring is created through mutation of the underlying latent vector $z$. Based on the results of~\cite{roziere2020evolgan}, 1/length of $z$ = 1/100 is chosen as individual mutation rate. If a gene is mutated, the mutation comprises the addition of a random number drawn from a standard normal distribution. Survivor selection then takes place on the basis of deterministic elitist replacement, meaning that only the better evaluated image is kept. The quality of the images resulting as phenotypes is automatically evaluated using NIMA~\cite{talebi2018nima}. In trade-off between quality increase and preservation of diversity, each local search spans 100 generations, since too many generations tend to result in less preserved diversity~\cite{roziere2020evolgan}.

\subsubsection{Preservation of Diversity}
Moreover, a metric is introduced that aims to keep the diversity in the population high. On the one hand, this has the goal of exploring the latent space quickly, since in IEC due to user fatigue usually only a small part of the search space is considered. On the other hand, this is intended to avoid user frustration, which can emerge after rating similar images over several generations and a feeling of being in a ``blind alley''~\cite{eiben2003EC}.

\begin{figure}[t]
    \centering
    \subfigure[]{\includegraphics[width=0.24\textwidth]{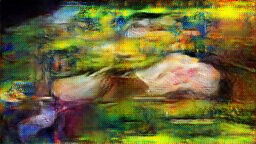}} 
    \subfigure[]{\includegraphics[width=0.24\textwidth]{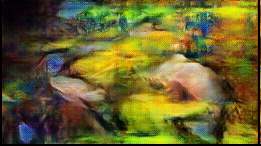}} 
    \subfigure[]{\includegraphics[width=0.232\textwidth]{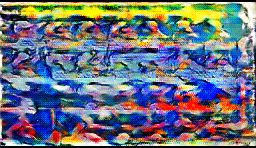}}
    \subfigure[]{\includegraphics[width=0.24\textwidth]{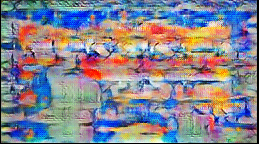}}
    \caption{Illustration of the \textit{preserve diversity} metric. While the examples (a) and (b) would be judged as too similar, the examples (c) and (d) exhibit sufficient differences.}
    \label{fig:Preserve_Div}
\end{figure}

The metric chosen is the sum of the absolute distances between individual genes, which is not allowed to fall below a certain threshold. This threshold is set to 25, and examples of the effects of this metric can be found in Figure~\ref{fig:Preserve_Div}. After crossover and mutation, this metric is applied to the entire population. Given that two images are too similar, one is replaced by a random immigrant, i.e. an image consisting of a completely new random latent vector. Random immigrants are a way to explicitly increase diversity~\cite{Immigrants}, and have also been used for this purpose in previous work on IEC in combination with GANs~\cite{bontrager2018deep,xin2021object}. In order to increase the quality of the random immigrants, they are first undergoing a local search over 100 generations, as implemented as mutation. 

\subsection{Evaluation}

The final evaluation by human participants of the results obtained with this framework consists of three parts. Each part is composed of pairwise comparisons, where two images are placed next to each other and participants have to decide which one they prefer. In the first part, 20 random images are compared before and after performing the local search to test its effect. In the second and third part, the results of the automatic and collaborative evolution are compared with random images as in the first generation. For this purpose, 10 images each were selected from the hall of fame, the overall best images during the evolutionary process, and randomly combined with the comparison images. Throughout the questionnaire, the order of the images is balanced to avoid position bias. An exemplary excerpt can be found in the supplementary material in Figure~\ref{fig:survey_final}. 

A total of $N = 31$ participants were recruited for the evaluation. Among them were 16 men, 13 women, and one person each who indicated a different gender or did not want to indicate their gender. The participants were between 20 and 87 years old ($M = 31.8, SD = 13.9$). In the evaluation, none of the participants involved in the collaborative evolution took part.

\section{Results}

\subsection{Local Search} \label{Results_LS}

\begin{figure}[h]
    \centering
    \subfigure[]{\includegraphics[width=0.40\textwidth]{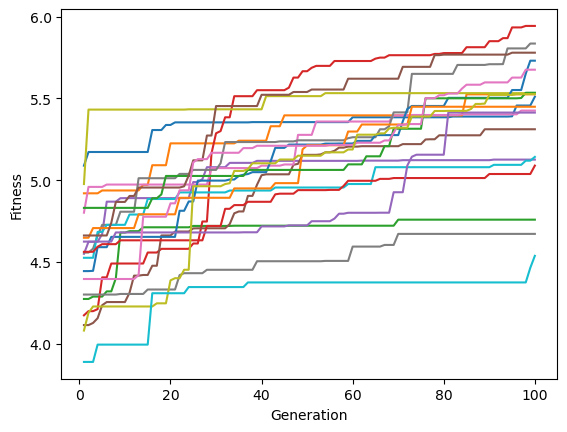}}
    \hspace{10mm}
    \subfigure[]{\includegraphics[width=0.27\textwidth]{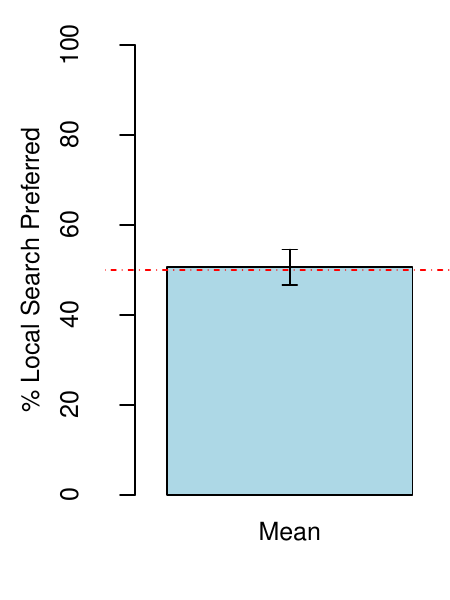}}
    \caption{(a) Fitness improvement of 20 random images over 100 generations of local search according to the automatic evaluation metric. (b) Proportion of participants preferring the local search results over the original image, averaged over all 20 images.}
    \label{fig:LS_results}
\end{figure}

Figure~\ref{fig:LS_results}(a) shows the fitness improvement of 20 random images that underwent local search for 100 generations as it was applied as mutation. These 20 images were also used to evaluate whether local search resulted in quality improvement for human participants. 

\begin{figure}[b]
    \centering
    \subfigure[]{\includegraphics[width=0.24\textwidth]{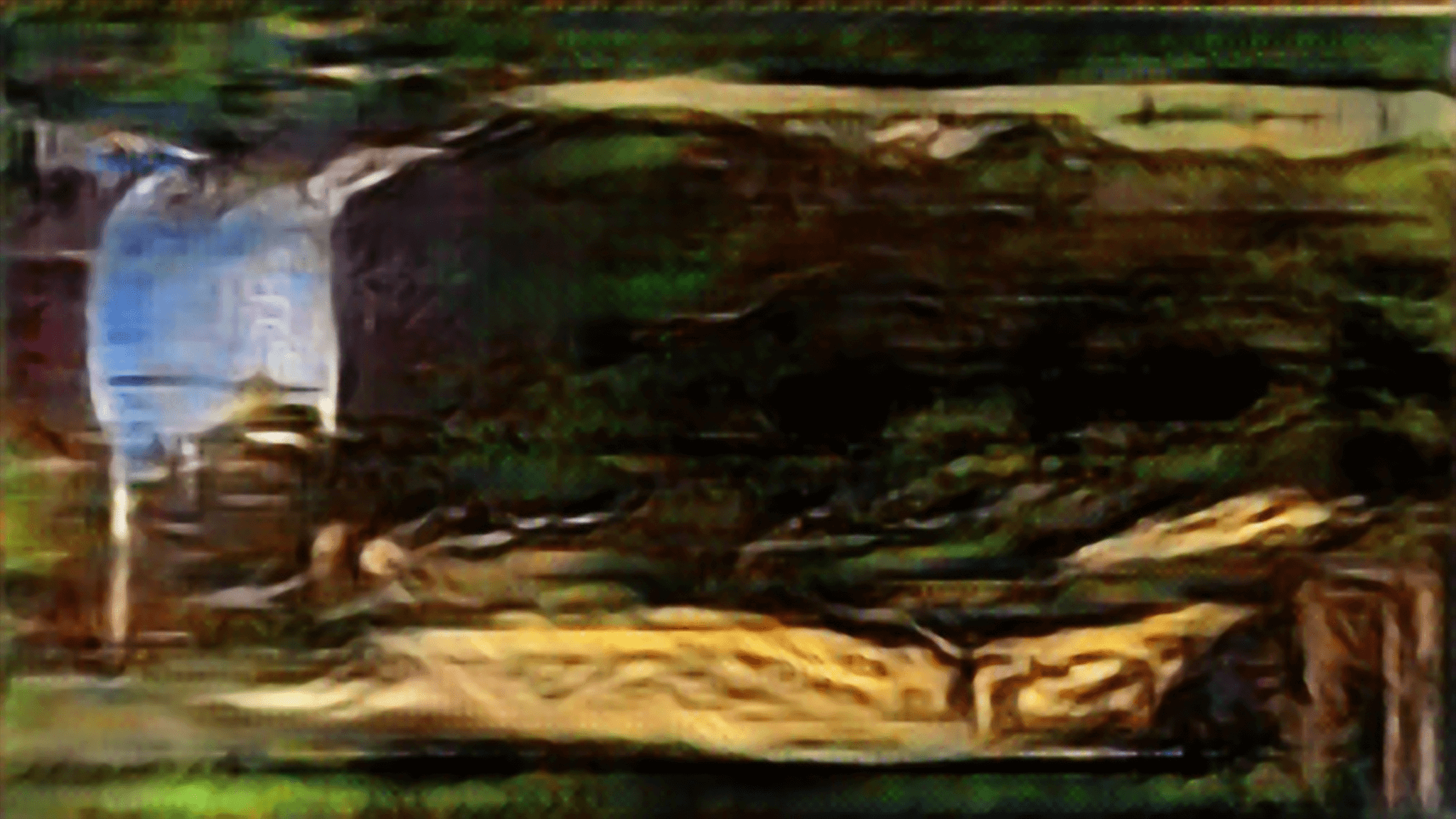}} 
    \subfigure[]{\includegraphics[width=0.24\textwidth]{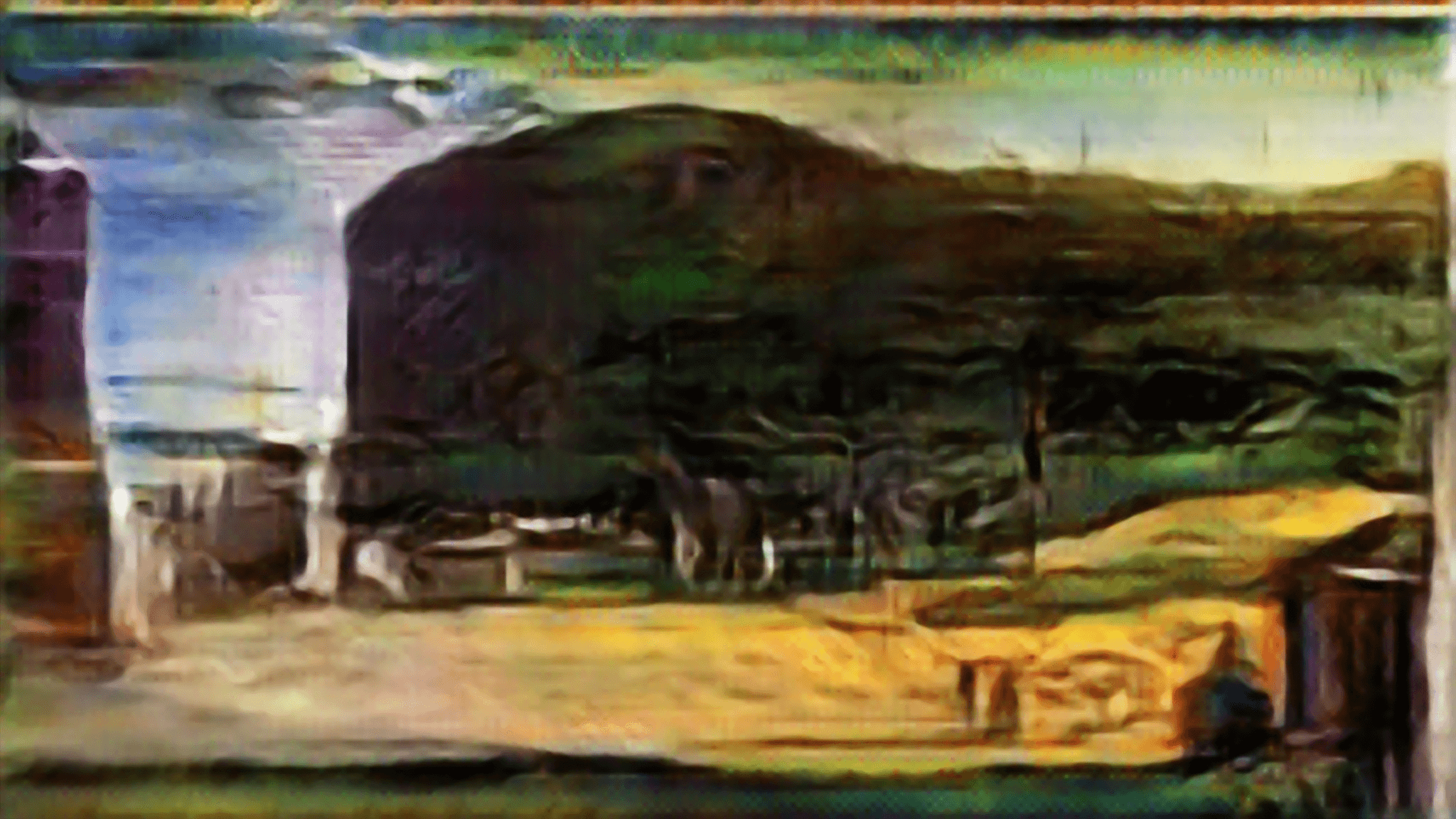}}
    \subfigure[]{\includegraphics[width=0.24\textwidth]{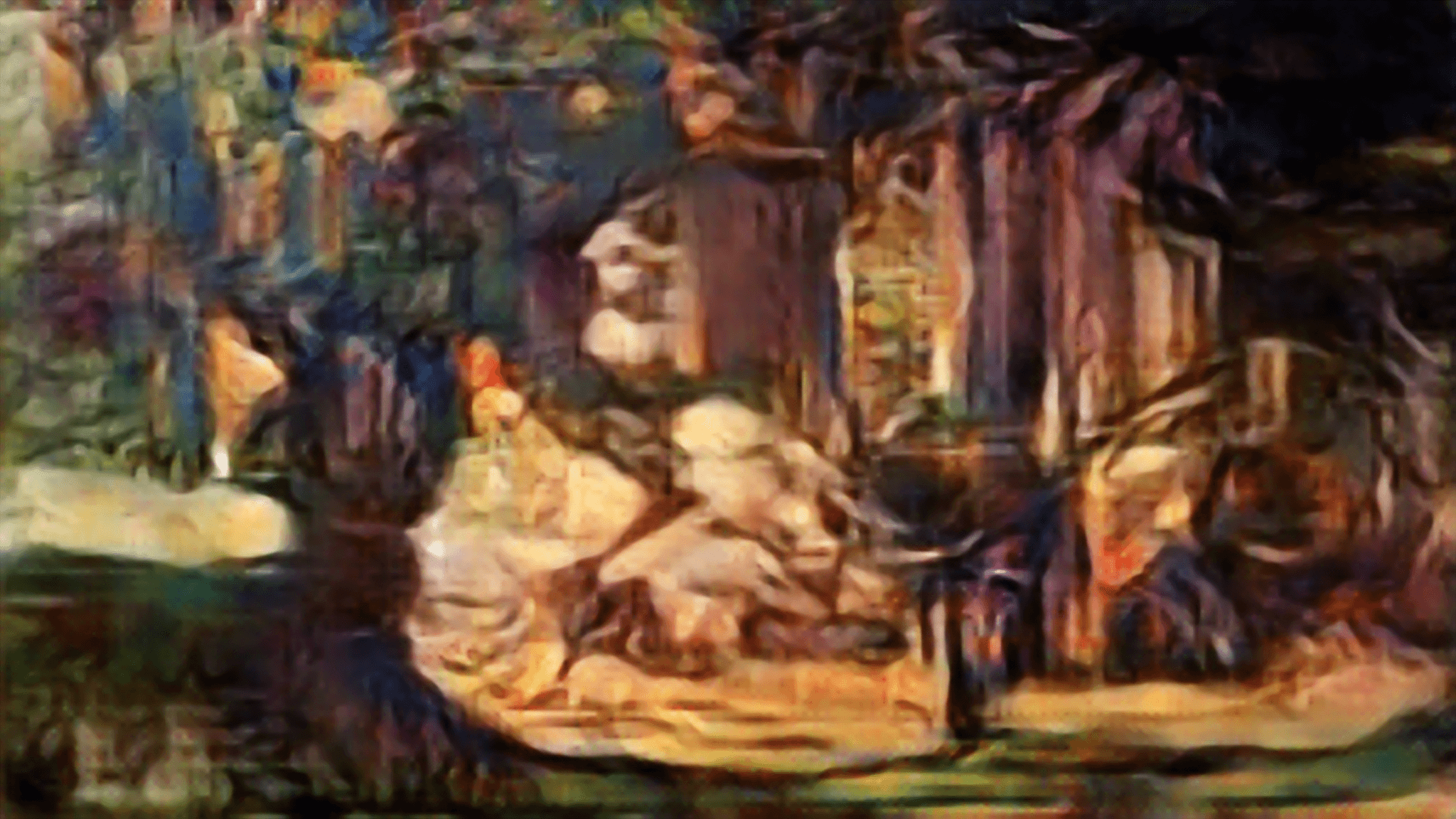}}
    \subfigure[]{\includegraphics[width=0.24\textwidth]{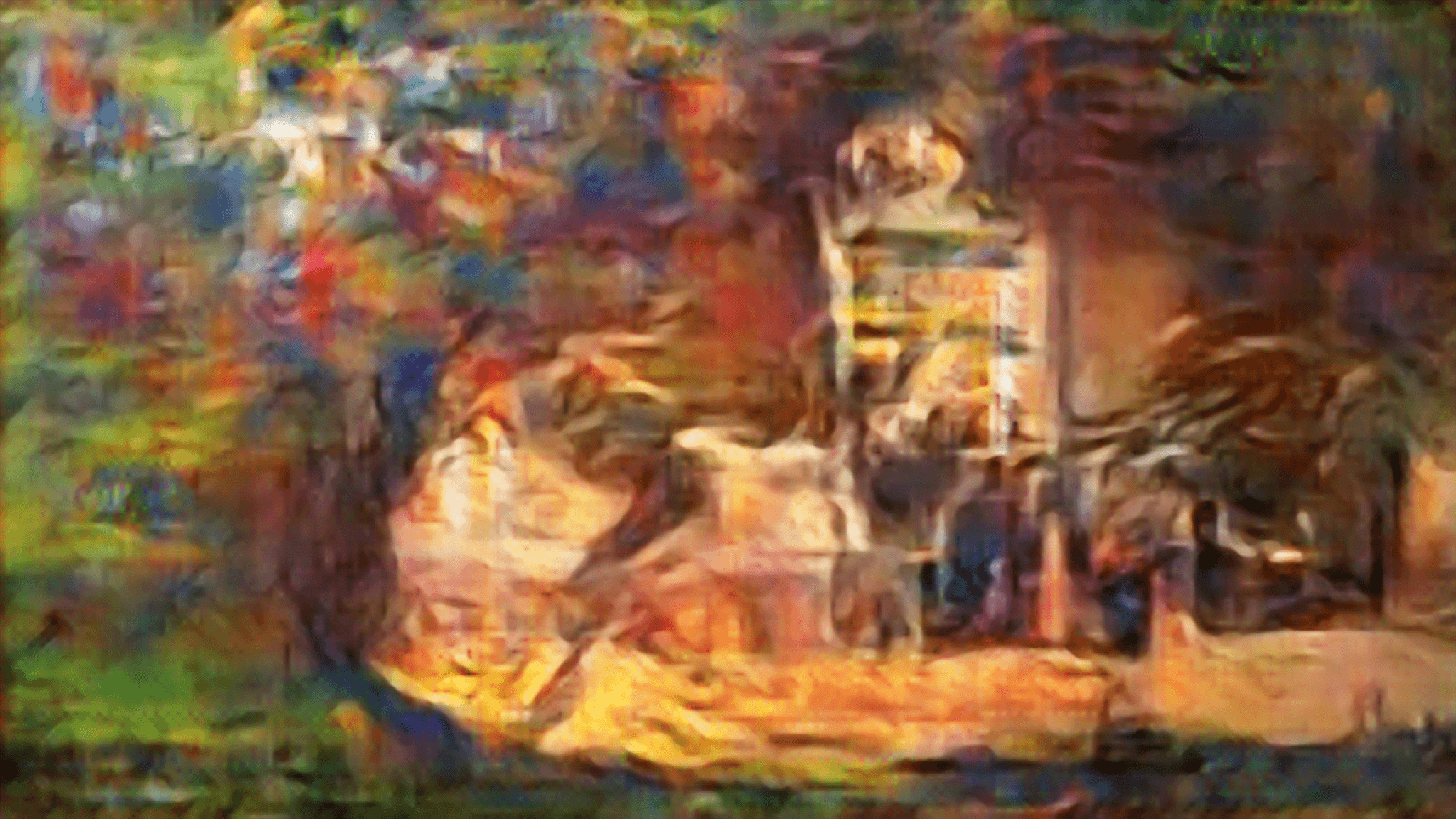}}
    \caption{Images (b) and (d) result from images (a) and (c) using the local search. (b) was preferred in 77\% over (a), while (d) was preferred only in 10\% over (c).}
    \label{fig:LS_Example}
\end{figure}

Figure~\ref{fig:LS_results}(b) illustrates the proportion at which the local search was preferred over the original image, averaged over all 20 images. The mean value of 51\% shows that the results are preferred only in slightly more than half of the cases, which corresponds to random level. Looking at the individual comparisons, besides many close decisions, local search led to clear improvements in some cases, but also to clear deteriorations in others. The most successful and least successful local searches are exemplified in Figure~\ref{fig:LS_Example}. Overall, it seems that sharper contours and more intense colours were perceived as improvements, while colour or composition changes received less approval.

\subsection{Automatic Evolution}

\begin{figure}[h]
    \centering
    \includegraphics[width=0.45\textwidth]{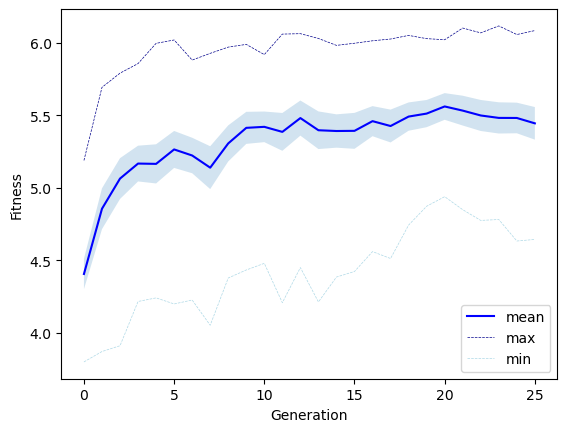}
    \caption{Mean fitness through generations of the automatic evolution, with the shaded area representing the standard error. The results are averaged over five runs.}
    \label{fig:AutomaticFitness}
\end{figure}

The evolution based on the automatic aesthetic evaluation metric resulted in increasing fitness through the generations. This trend can be seen in Figure~\ref{fig:AutomaticFitness}, which displays results averaged over five runs. The fitness increased particularly at the beginning and seemed to reach a plateau at the end. This trend is presumably supported by the fact that the mutation used also evolved the individual images with respect to the same aesthetic evaluation metric. 

\begin{figure}[h]
    \centering
    \subfigure[]{\includegraphics[width=0.32\textwidth]{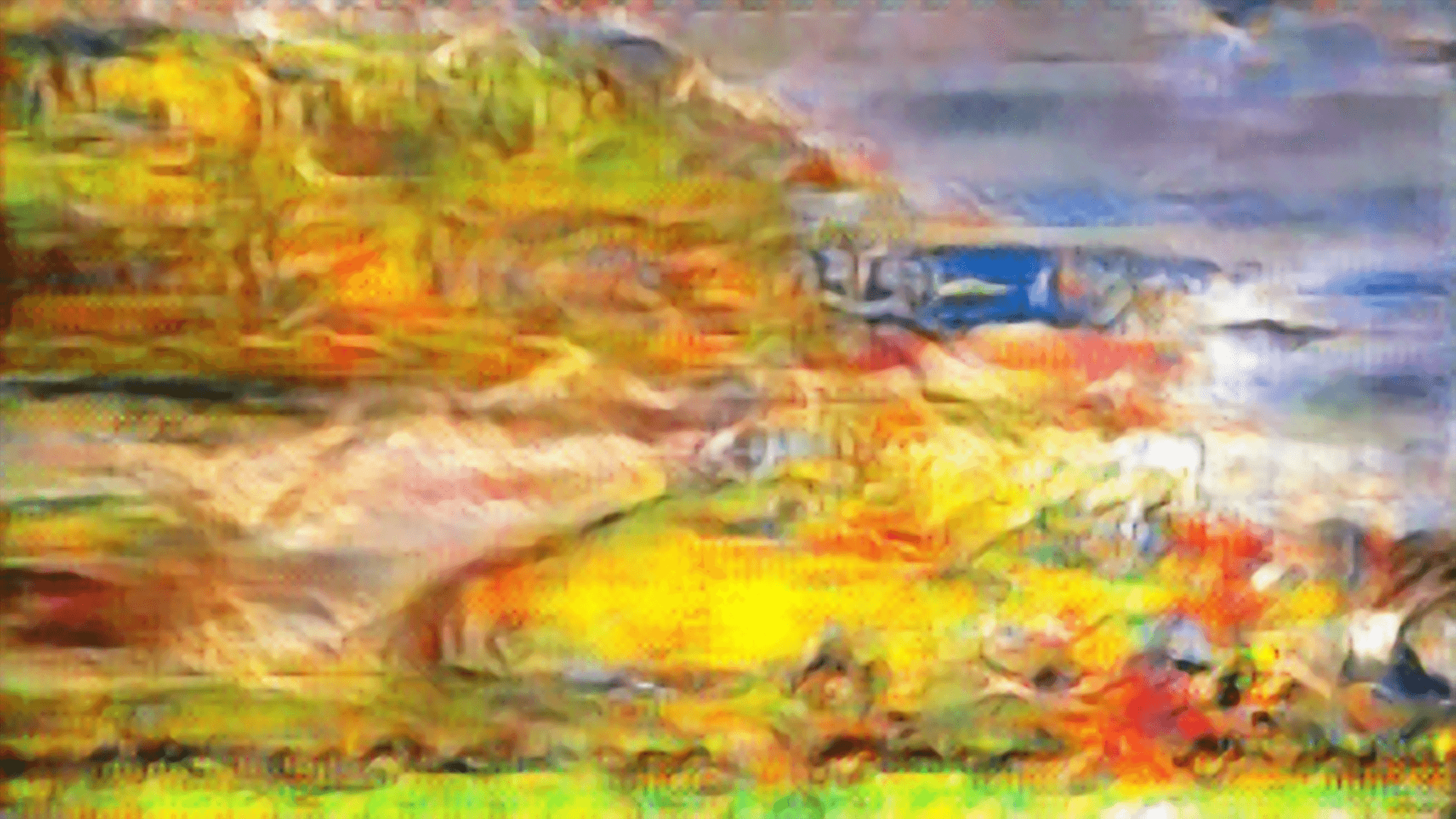}} 
    \subfigure[]{\includegraphics[width=0.32\textwidth]{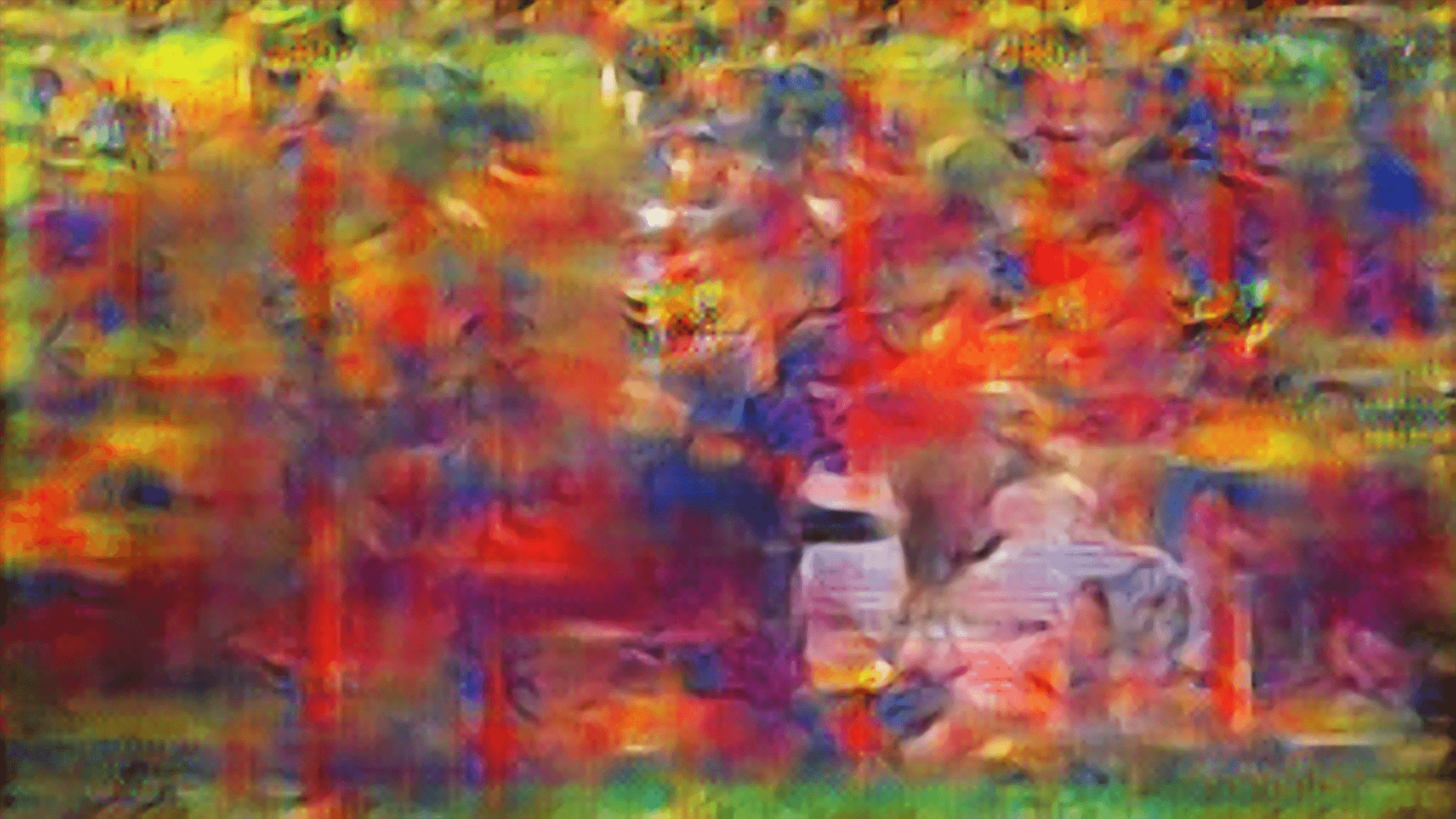}}
    \subfigure[]{\includegraphics[width=0.32\textwidth]{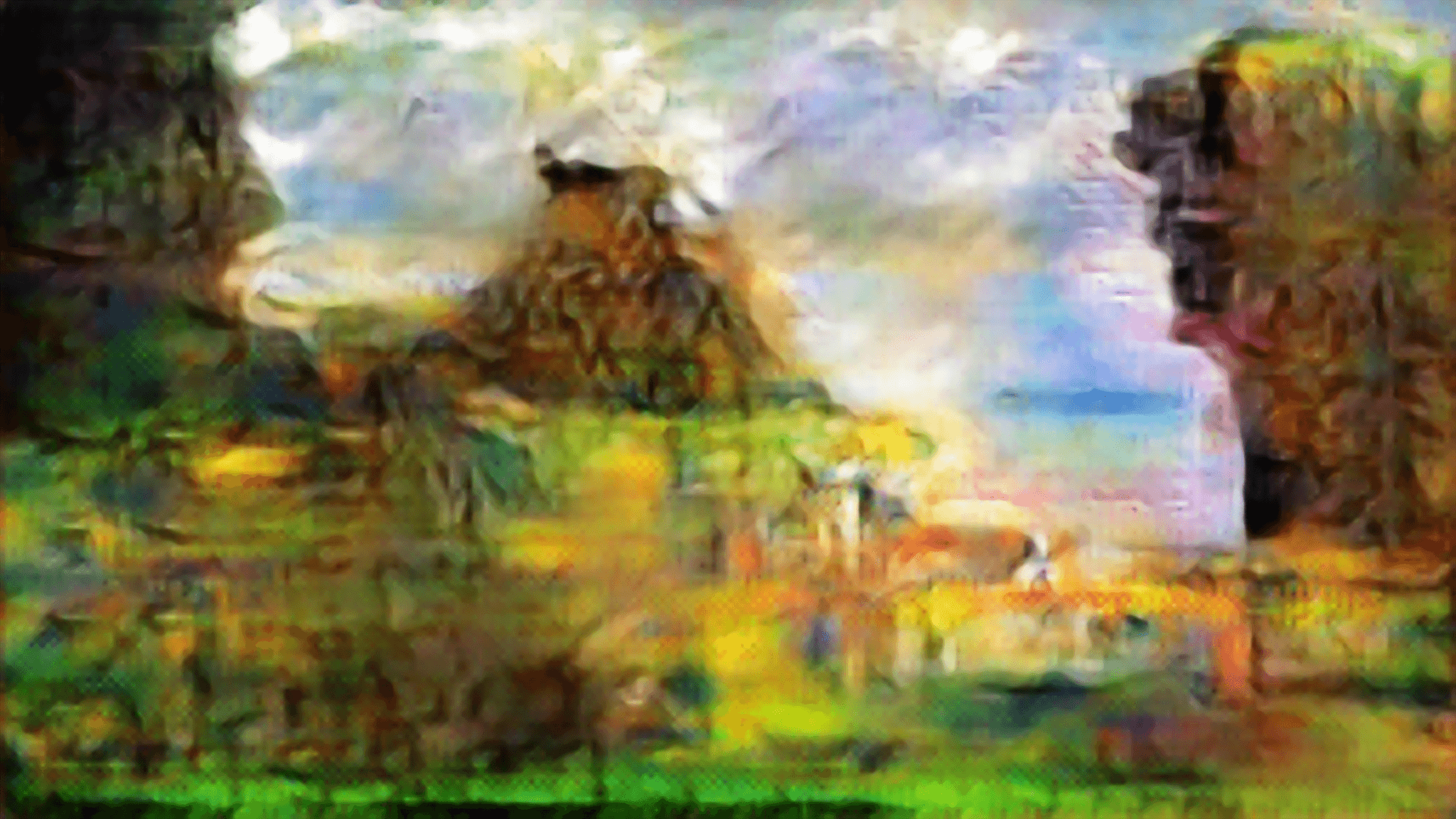}}
    \subfigure[]{\includegraphics[width=0.32\textwidth]{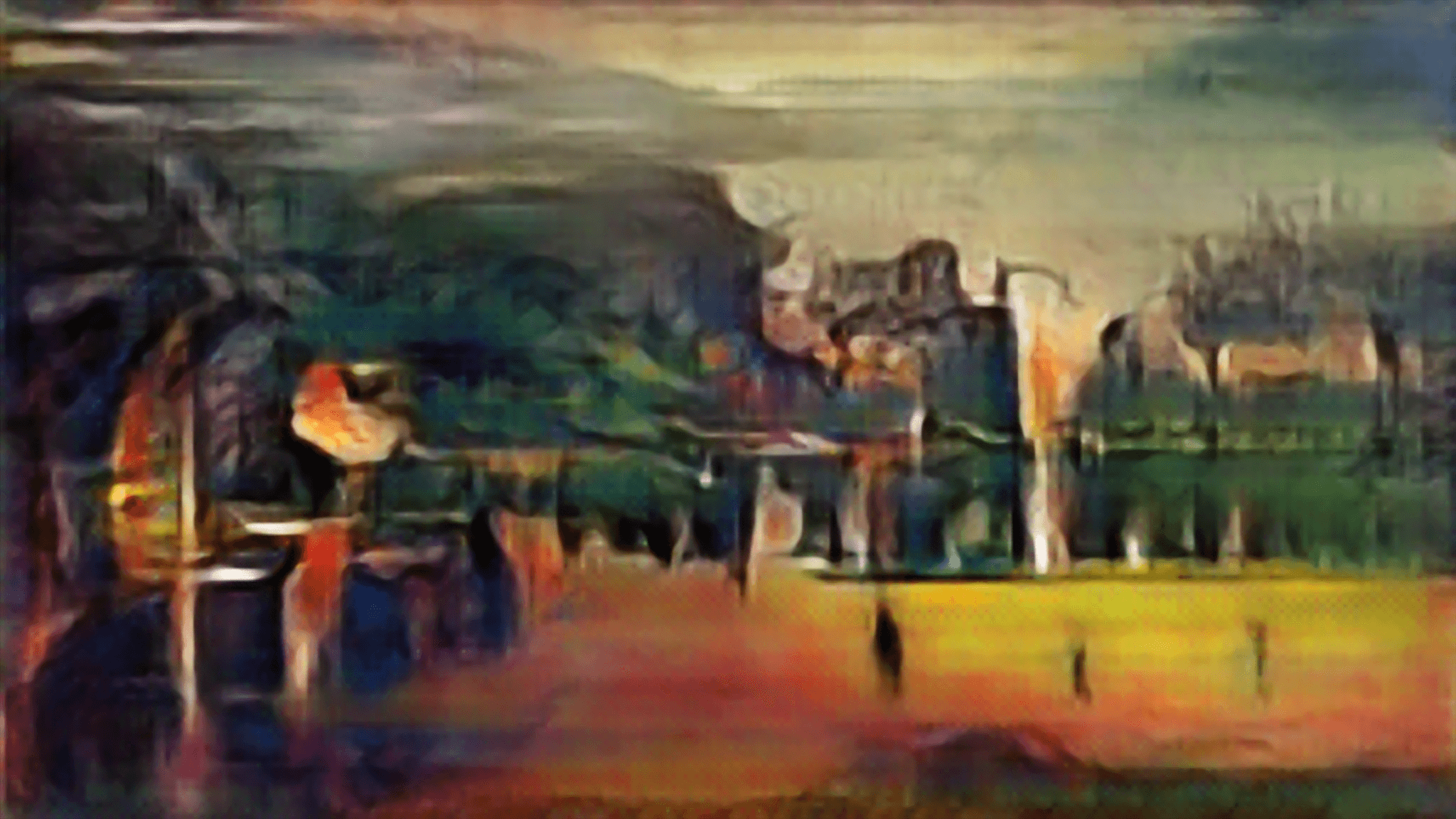}}
    \subfigure[]{\includegraphics[width=0.32\textwidth]{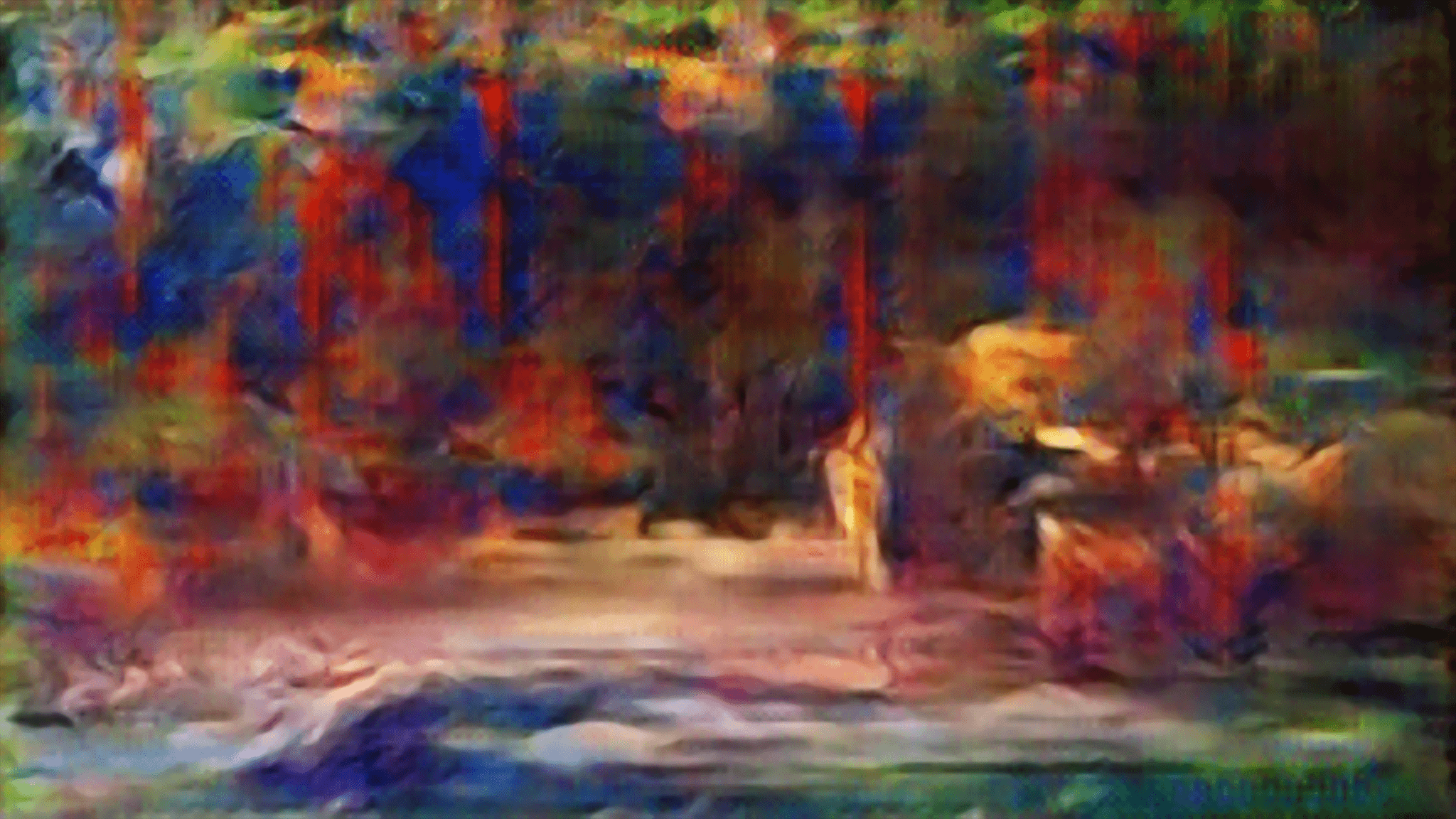}}
    \subfigure[]{\includegraphics[width=0.32\textwidth]{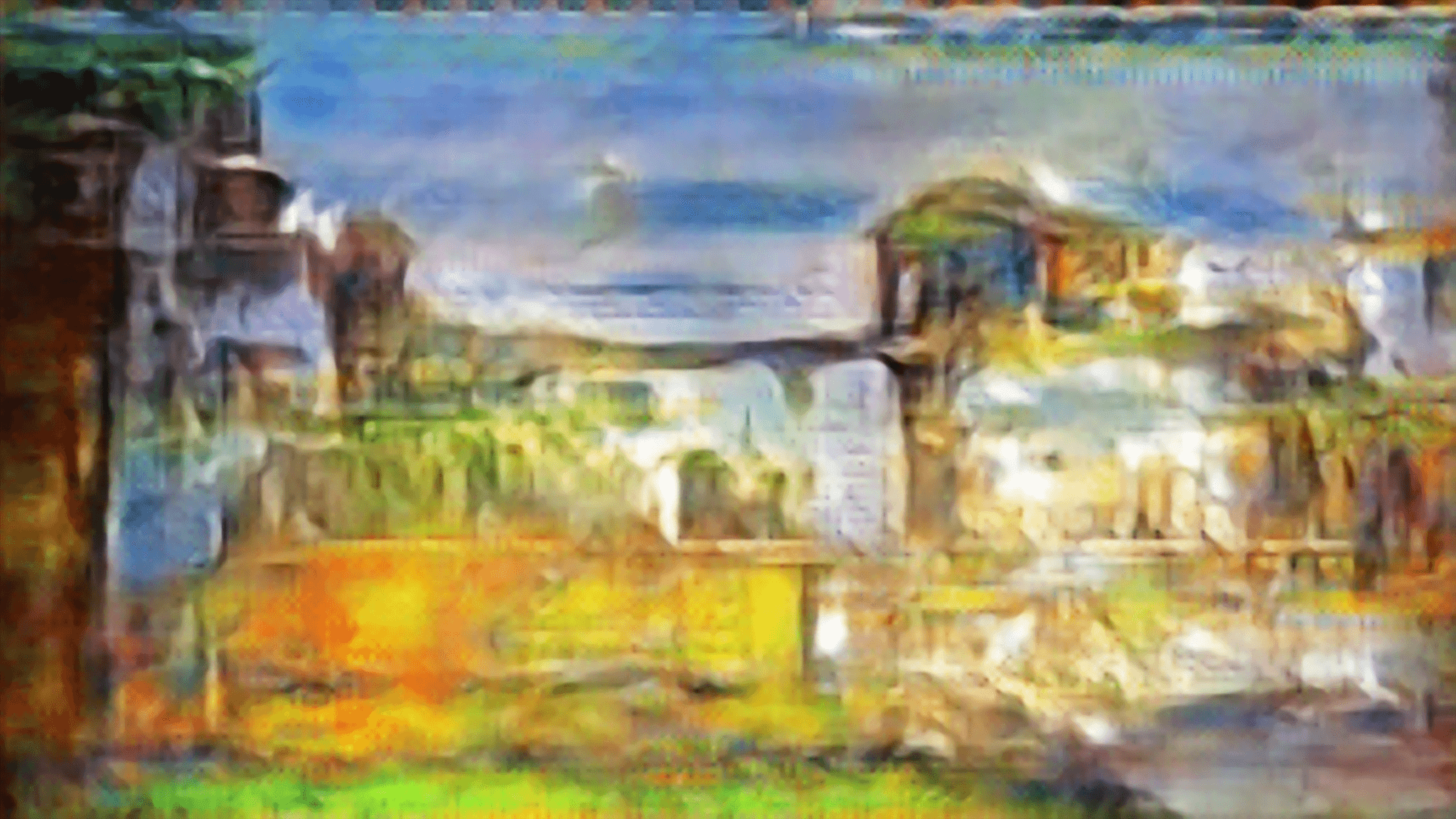}}
    \caption{A selection of the overall best images of the automatic evolution.}
    \label{fig:AutomaticEvolution_HOF}
\end{figure}

All images of the first and last generation are shown in the supplementary material in Figure~\ref{fig:Evolution}, a visualisation of the entire evolution can be found online\footnote{Visualisation of the automatic evolution: \textcolor{blue}{\url{https://youtu.be/JCRx3Ih_0hA}}}. A selection of the best images overall is displayed in Figure~\ref{fig:AutomaticEvolution_HOF}. It is apparent that the automatic evolution evolved in the direction of rather blurred images with less clear shapes. With the exception of the image in Figure~\ref{fig:AutomaticEvolution_HOF}(d), which has clearer structures and could be reminiscent of a landscape, diffuse contours predominate. Partly, a sky might be recognisable.

\subsubsection{Evaluation}
The evaluation regarding the attractiveness of the images obtained in the automatic evolution compared to random images as in the first generation showed that human participants preferred the obtained images on average in 49\% of the cases, which corresponds to random level. All individual comparisons and the resulting mean can be found in the supplementary material in Figure~\ref{fig:BigEvaluation}(a). Only two comparisons were clearly in favour of the automatic evolution results, while a majority of comparisons tended towards the random images.

\subsection{Collaborative Evolution}
\begin{figure}[h]
    \centering
    \subfigure[]{\includegraphics[width=0.45\textwidth]{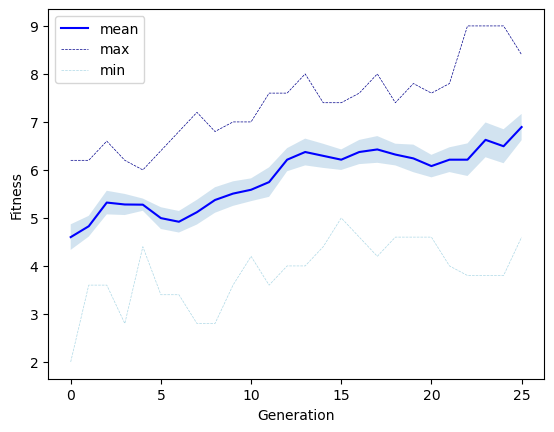}} 
    \hspace{3mm}
    \subfigure[]{\includegraphics[width=0.45\textwidth]{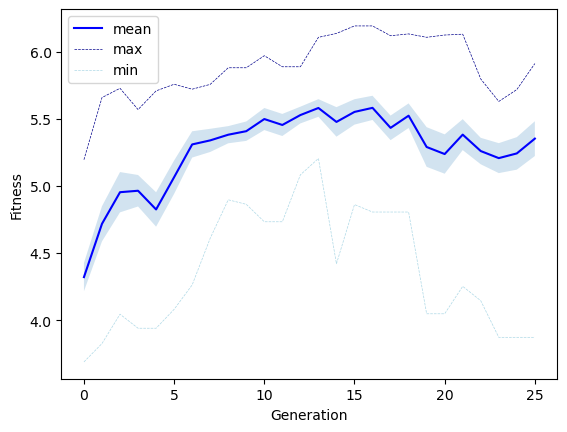}}
    \caption{(a) Mean fitness through generations of the collaborative interactive evolution. (b) The collaborative interactive evolution assessed by the automatic aesthetic evaluation metric. The shaded areas represent the standard error.}
    \label{fig:EvolutionInteractive_Results}
\end{figure}

The collaborative interactive evolution resulted in increasing fitness through the generations, too. This trend can be seen in Figure~\ref{fig:EvolutionInteractive_Results}(a). In contrast to the automatic evolution, this increase was rather linear and kept rising until the end. Figure~\ref{fig:EvolutionInteractive_Results}(b) shows how the automatic aesthetic evaluation metric would have assessed the fitness of the collaborative interactive evolution. Interestingly, similar to the actual automatic evolution in Figure~\ref{fig:AutomaticFitness}, the fitness is strongly increasing at the beginning and rather stagnant afterwards, which may also be due to the effect of the local search. Compared to the collaborative interactive evolution, there are similarities such as the local minima at generation 20, but the trends also reveal many differences. 

A total of 17 random immigrants were introduced into the population during the evolutionary process. Only in generation two, two immigrants were inserted, otherwise at most one. Interestingly, the immigrants were rated above average in 14 out of 17 cases and were on average 0.67 above the generation means. This could indicate that novelty was perceived as positive by human participants throughout the evolutionary process.

\begin{figure}[h]
    \centering
    \subfigure[]{\includegraphics[width=0.35\textwidth]{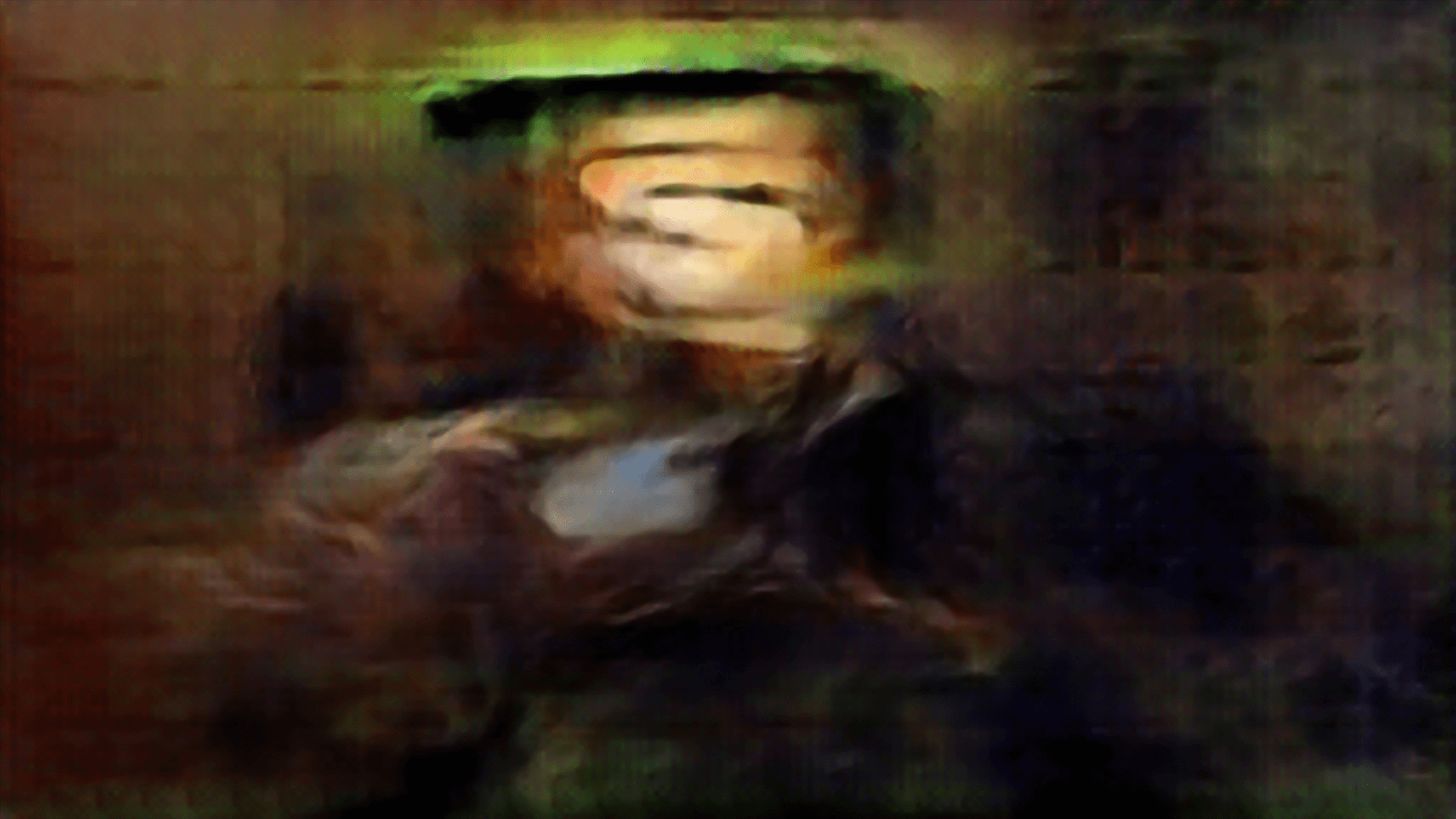}}
    \hspace{10mm}
    \subfigure[]{\includegraphics[width=0.35\textwidth]{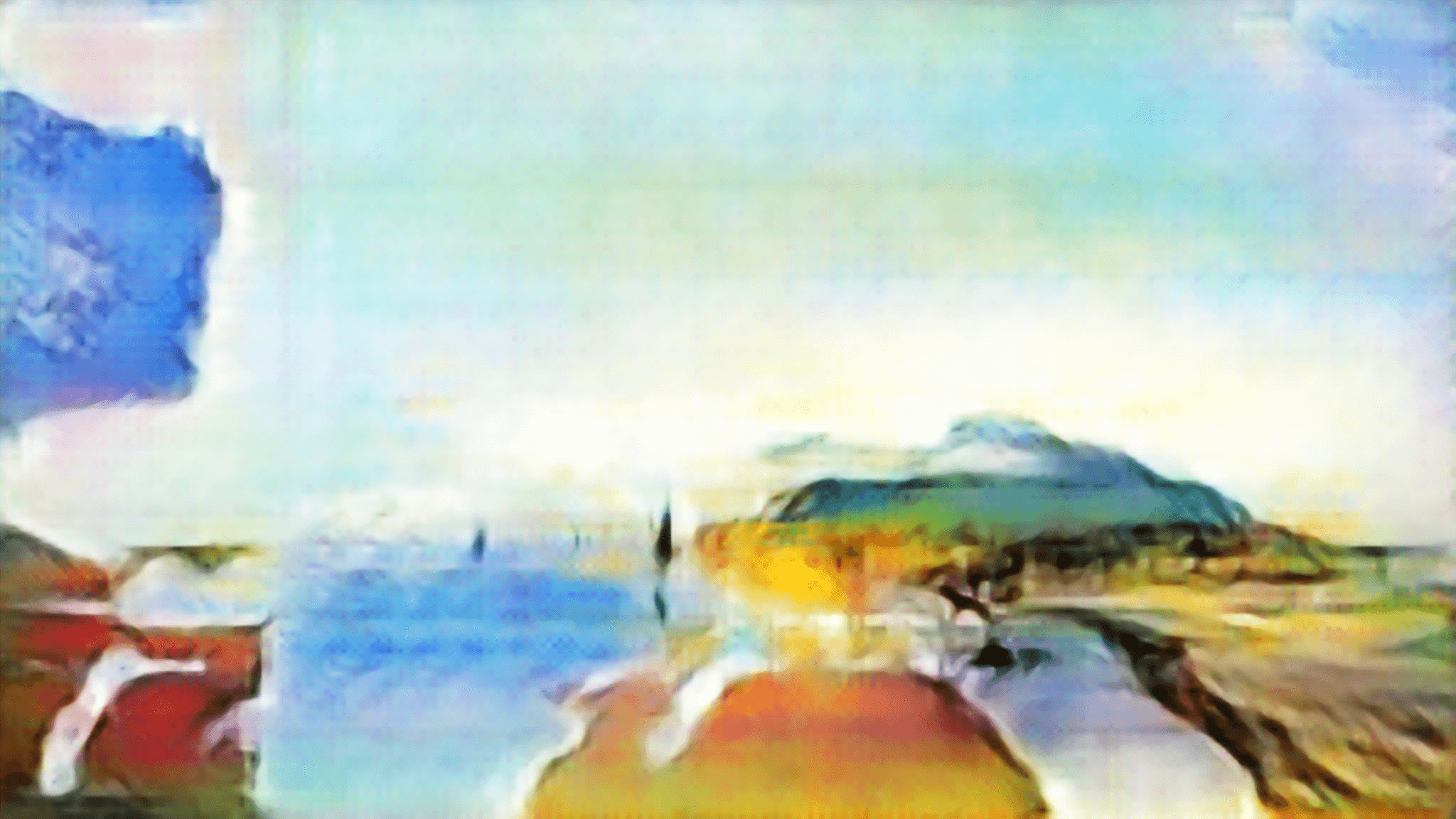}}
    \caption{(a) The image with the widest range in ratings (1-10). (b) The image with the closest agreement (all 8).}
    \label{fig:EvolutionInteractive_Subjectivity}
\end{figure}

The collaborative interactive approach further revealed how subjectively art images are perceived. The ratings differed significantly in some cases, with ranges across the entire scale (1-10) and standard deviations of up to $SD = 3.38$. For some images, however, the ratings were also highly similar. Figure~\ref{fig:EvolutionInteractive_Subjectivity} shows in (a) the image with the widest range in ratings and in (b) the image with the closest agreement in ratings. In fact, the image was the only one rated equally by all participants with an 8. 

\begin{figure}[h]
    \centering
    \subfigure[]{\includegraphics[width=0.32\textwidth]{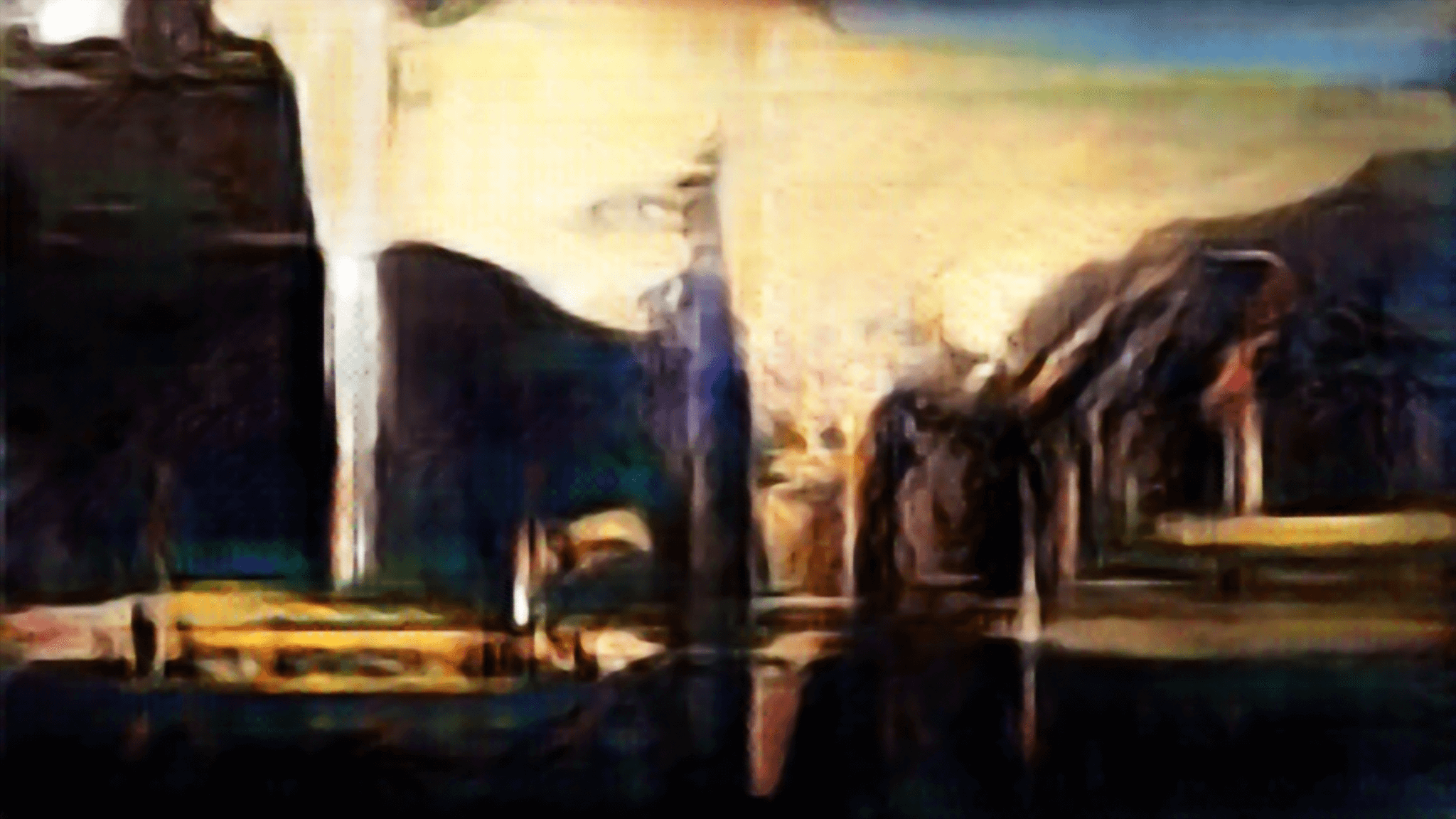}} 
    \subfigure[]{\includegraphics[width=0.32\textwidth]{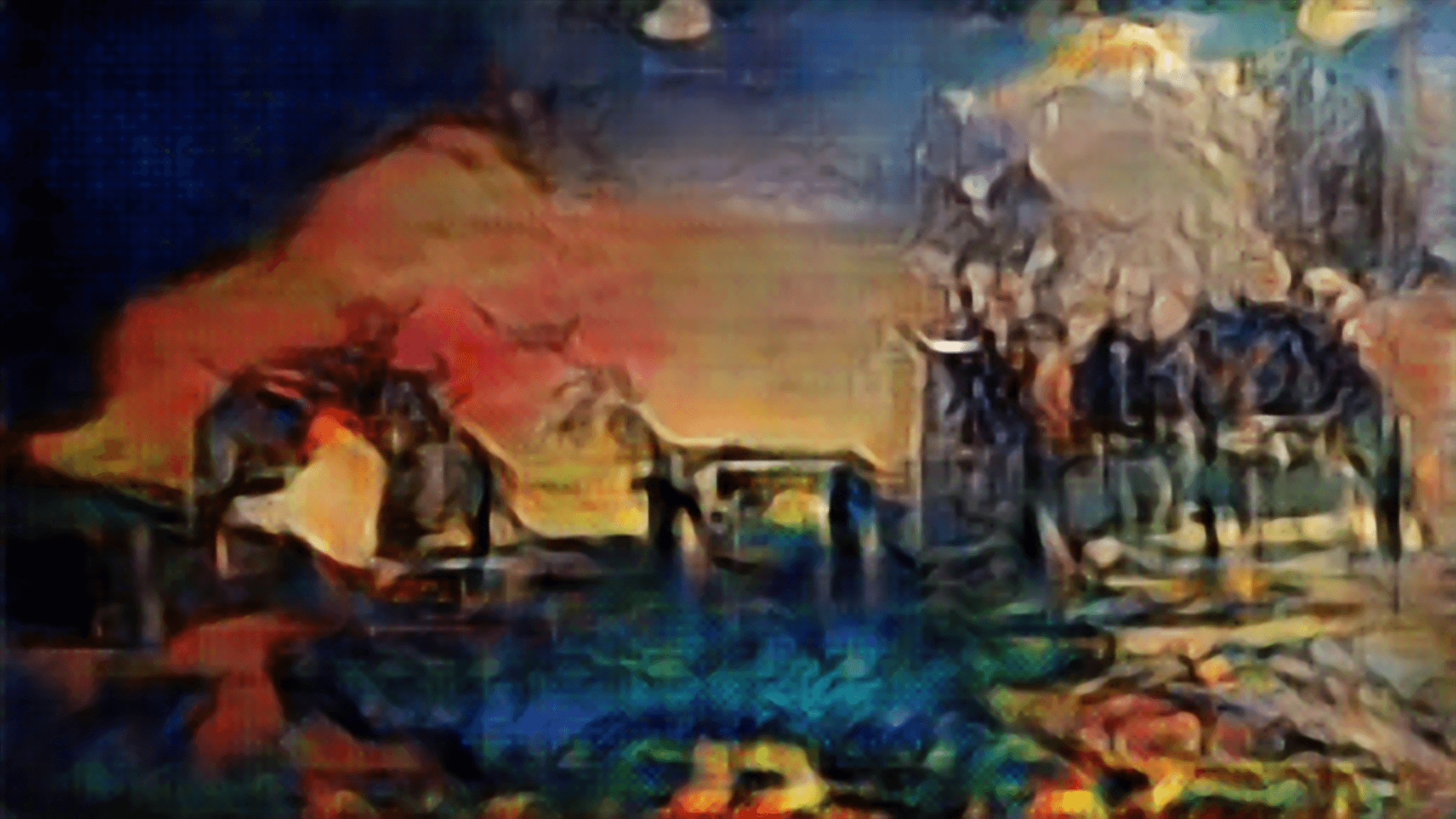}} 
    \subfigure[]{\includegraphics[width=0.32\textwidth]{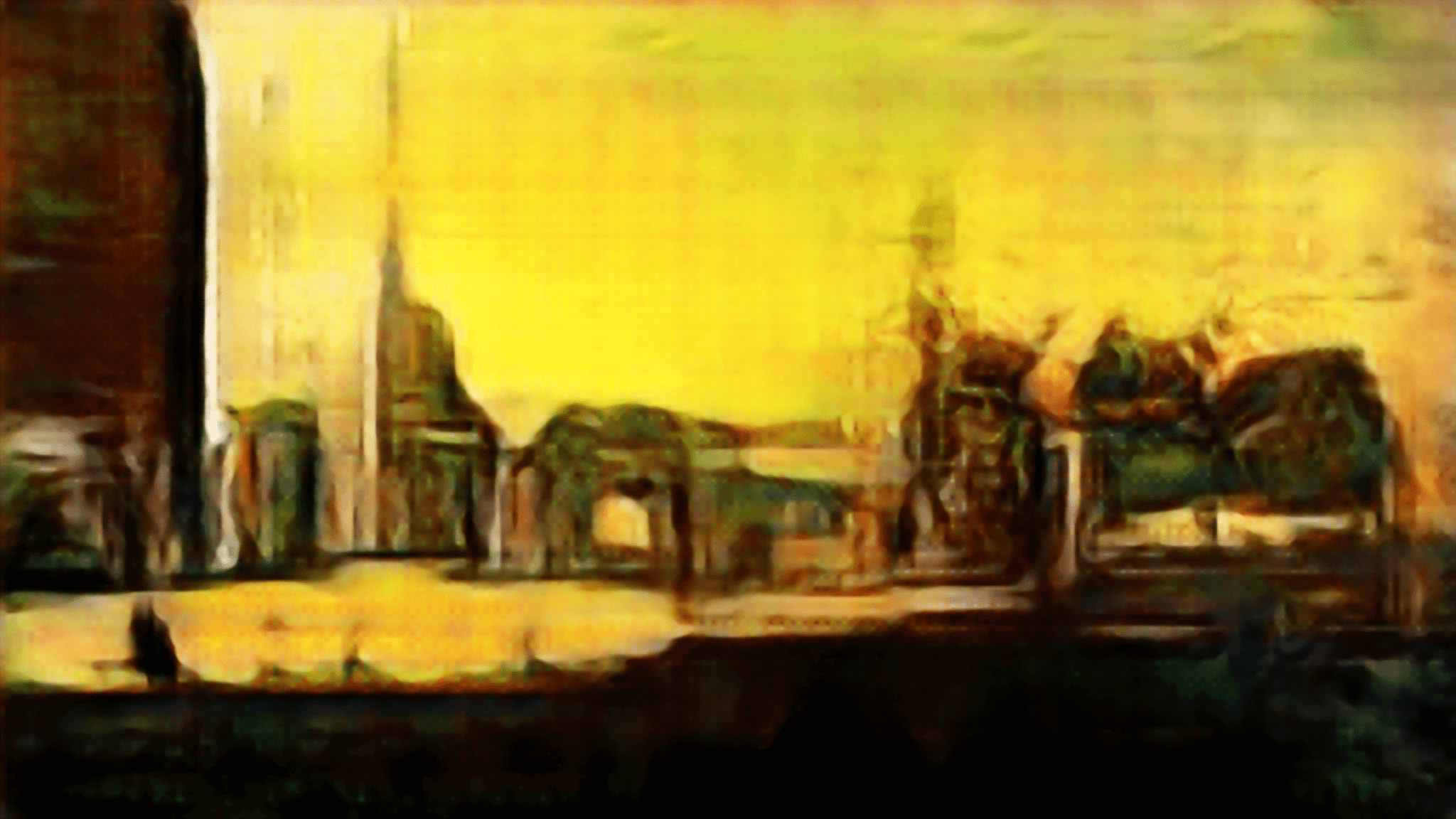}} 
    \subfigure[]{\includegraphics[width=0.32\textwidth]{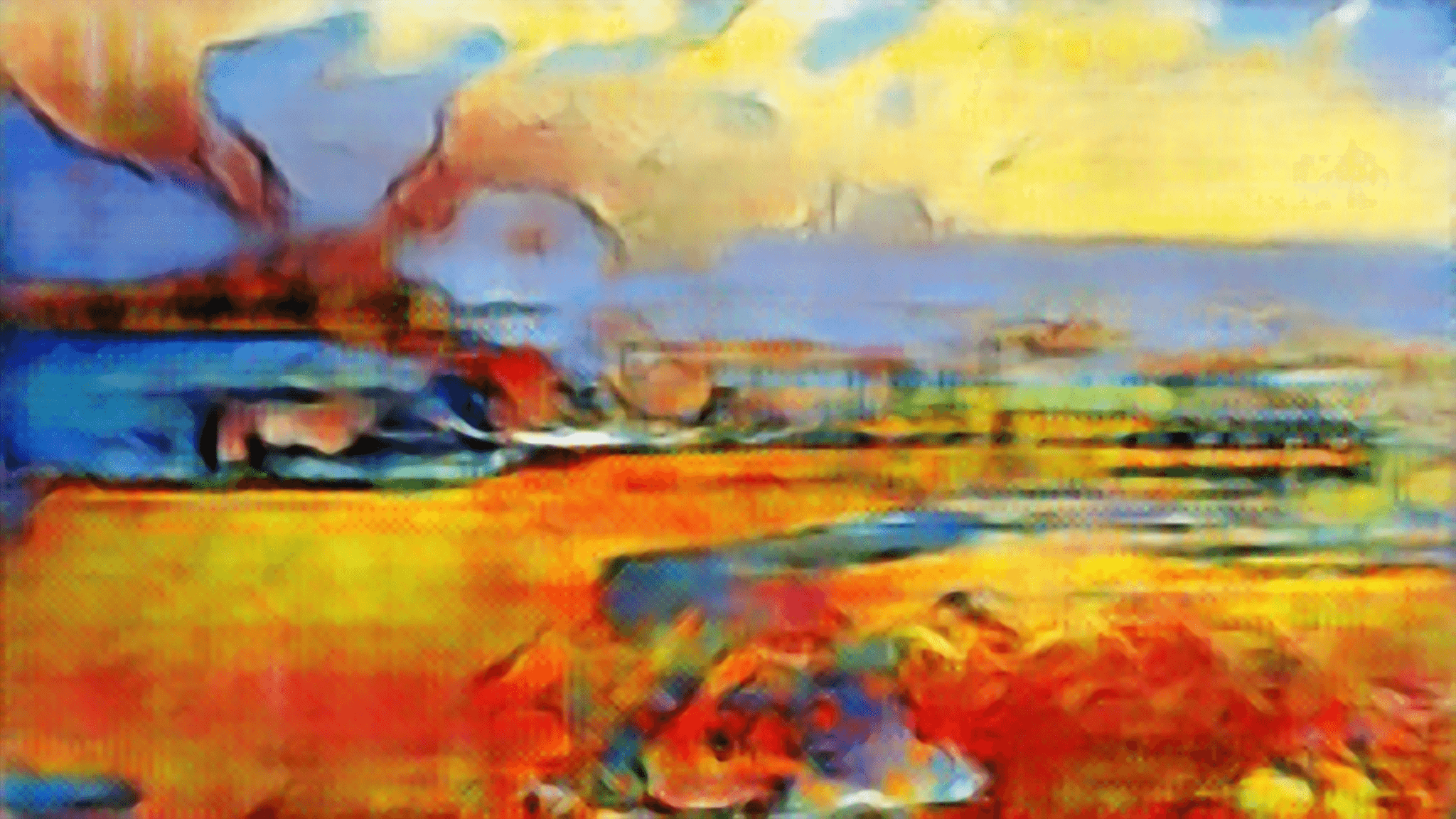}} 
    \subfigure[]{\includegraphics[width=0.32\textwidth]{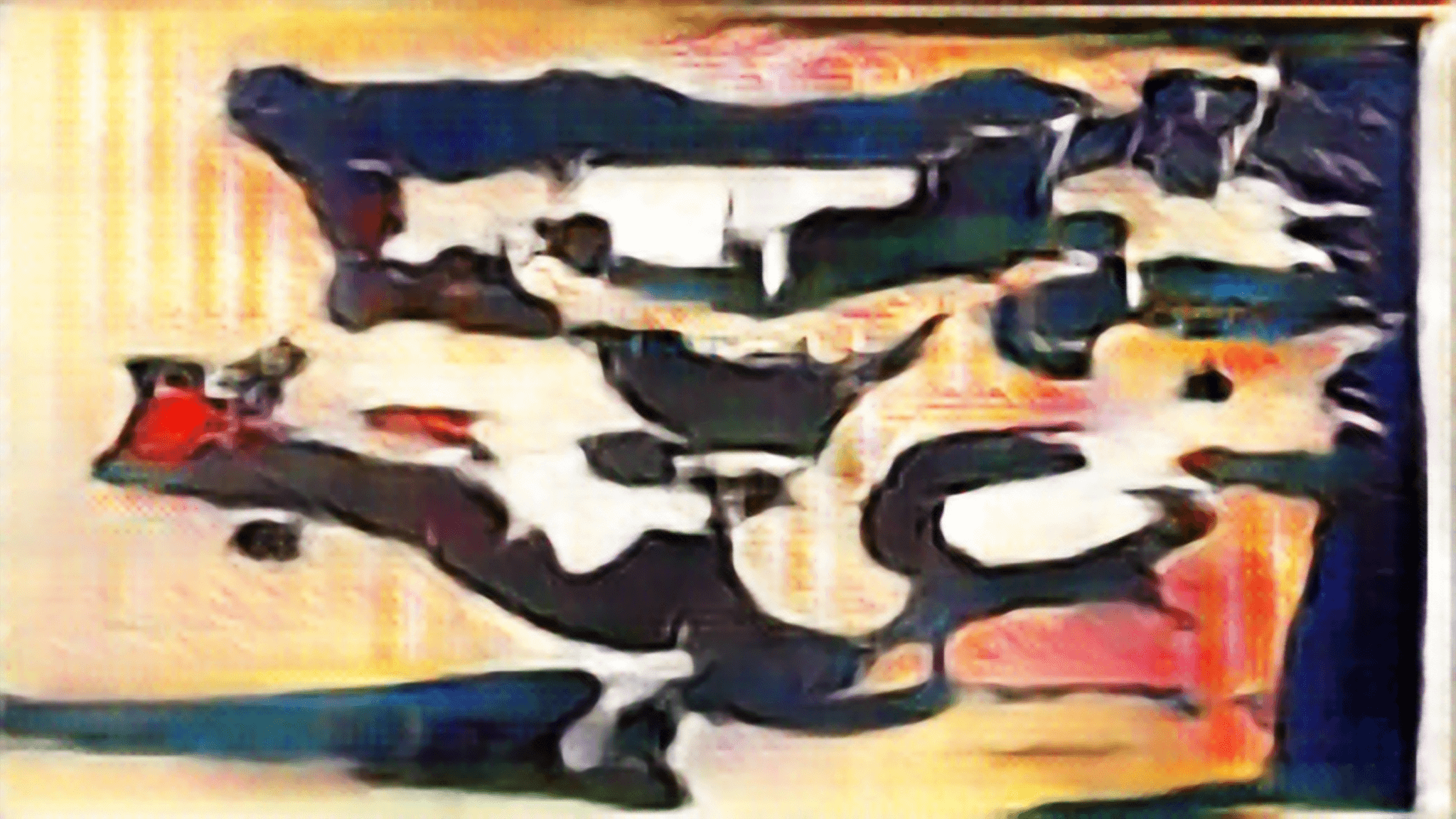}} 
    \subfigure[]{\includegraphics[width=0.32\textwidth]{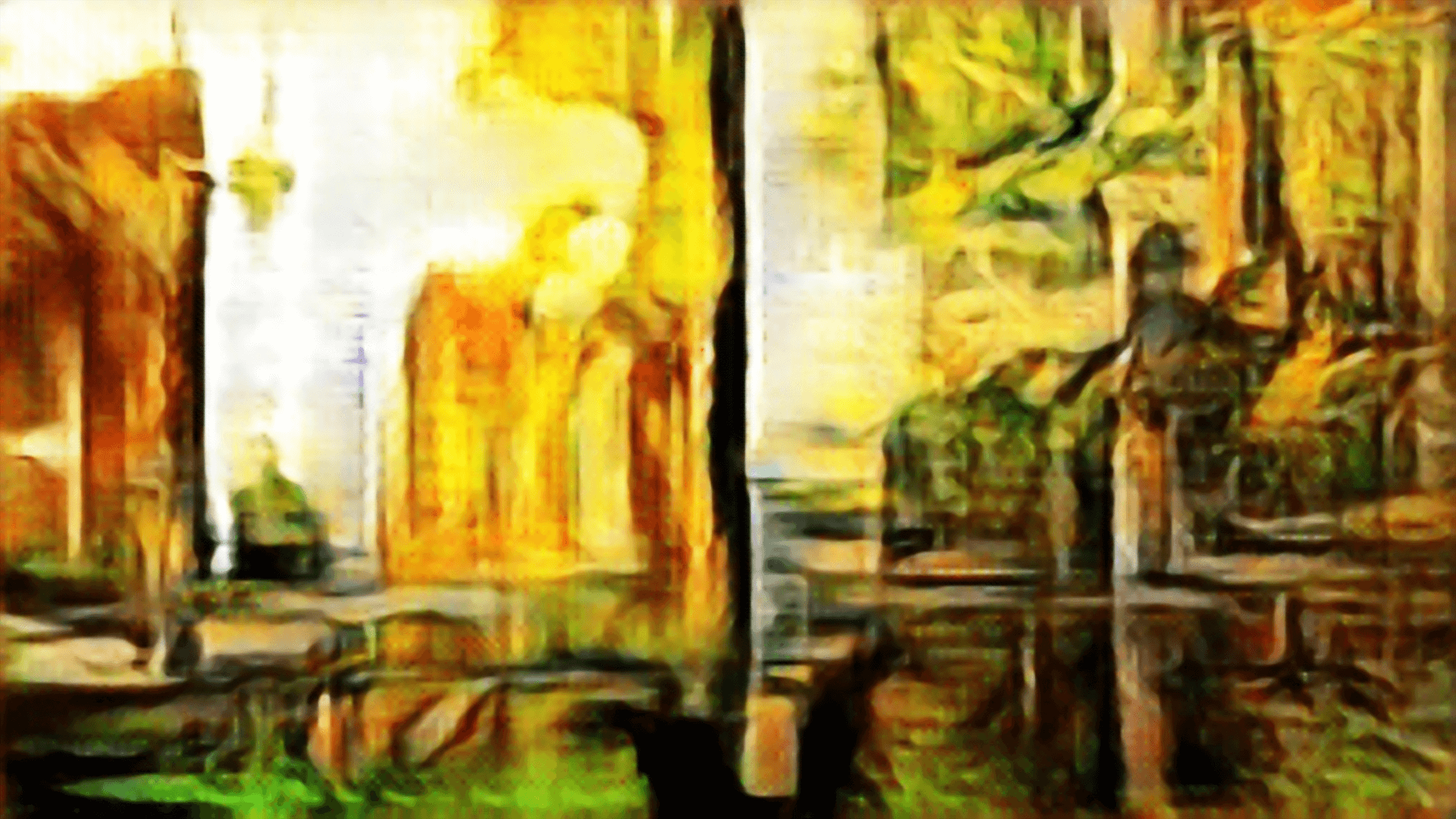}} 
    \caption{A selection of the overall best images of the collaborative evolution.}
    \label{fig:InteractiveEvolution_HOF}
\end{figure}

As with the automatic evolution, all images of the first and last generation are shown in the supplementary material in Figure~\ref{fig:Evolution}, and a visualisation of the entire evolution can be found online\footnote{Visualisation of the collaborative evolution: \textcolor{blue}{\url{https://youtu.be/rG_pLiX_UFo}}}. A selection of the best images overall is displayed in Figure~\ref{fig:InteractiveEvolution_HOF}. In contrast to the automatic evolution, sharper contours prevail. The images are less blurry and the colours appear more diverse, also the images seem to differ more from each other. Many of the images evoke landscape-like associations or are reminiscent of abstract art.

\subsubsection{Evaluation}
The evaluation regarding the attractiveness of the images obtained in the collaborative interactive evolution compared to random images as in the first generation showed that human participants preferred the obtained images on average in 60\% of the cases. Averaged across all ten comparisons and all assessments, an exact binomial test revealed that the results of the collaborative evolution were significantly preferred over random images ($T = 185, p < .001$). All individual comparisons and the resulting mean can be found in the supplementary material in Figure~\ref{fig:BigEvaluation}(b). Some decisions were close and do not point clearly in one direction considering the standard errors. However, with descriptively eight decisions in favour of the results of the collaborative evolution and in at least three cases a strong preference, a clear tendency is recognisable.

\section{Conclusion}

In this work, we introduced a novel framework to explore the latent space of possible art images in a GAN using evolutionary computing. It was shown that the developed framework led to an increase in quality over the evolutionary process, using both a collaborative interactive and an automatic aesthetic evaluation metric. The evolutionary algorithm was hybrid, since a local search based on an automatic evaluation metric was incorporated as an intelligent mutation. However, the evaluation by human participants revealed that the local search did not lead to improvements beyond a random level. Furthermore, only the results of the collaborative interactive evolution, but not of the automatic aesthetic evolution were found to be significantly more attractive than randomly generated images. This highlights that automatic aesthetic evaluation of art is challenging and emphasises the importance of human guidance in the evolution of art.

It was demonstrated that the use of the generator part of a GAN as genotype-to-phenotype mapping offers a promising approach for the evolution of art. Throughout all generations, diverse images were generated that can be considered as art images, and the generated images increased in their attractiveness over time. While the already high quality of the randomly generated images makes it more difficult to create significantly more attractive ones in the evolutionary process, it benefits the ongoing interest of users and the prevention of user fatigue. Further, it allows techniques such as random immigrants to be used without leading to substantial fitness loss, and is thus also conducive to accelerated exploration of latent space. The introduced collaborative interactive approach indicated its usefulness due to the subjectivity of art and the positive outcomes, and opens up a multitude of future research possibilities.

\newpage
%
%
\bibliographystyle{splncs04}
\bibliography{bibliography}

\newpage
\section*{Supplementary Material}

\begin{figure}[h]
    \centering
    \includegraphics[width=0.87\textwidth]{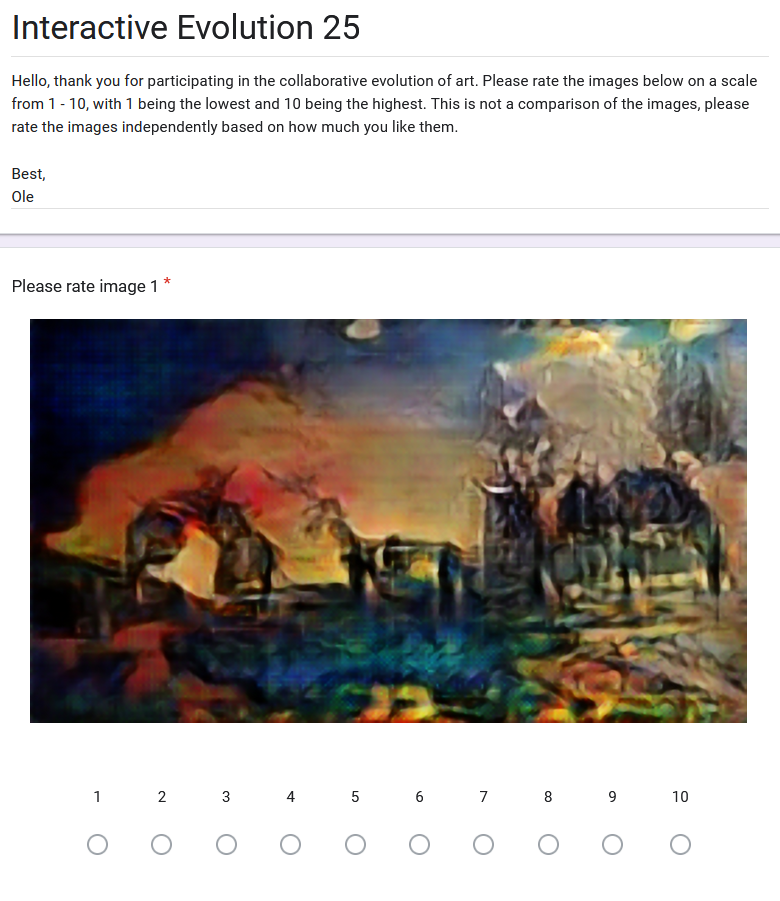}
    \caption{Illustrative excerpt of the collaborative interactive evolution questionnaires.}
    \label{fig:survey_coll}
\end{figure}

\newpage

\begin{figure}[h]
    \centering
    \includegraphics[width=0.99\textwidth]{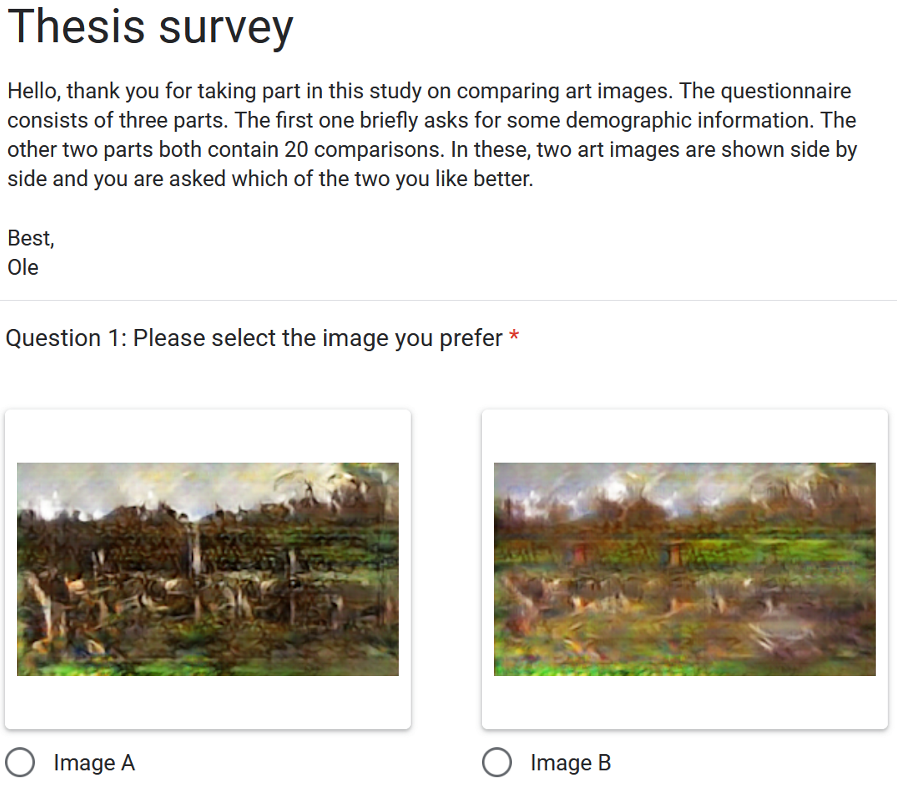}
    \caption{Illustrative excerpt of the final evaluation questionnaire.}
    \label{fig:survey_final}
\end{figure}

\newpage

\begin{figure}[h]
    \centering
    \subfigure[]{\includegraphics[width=0.99\textwidth]{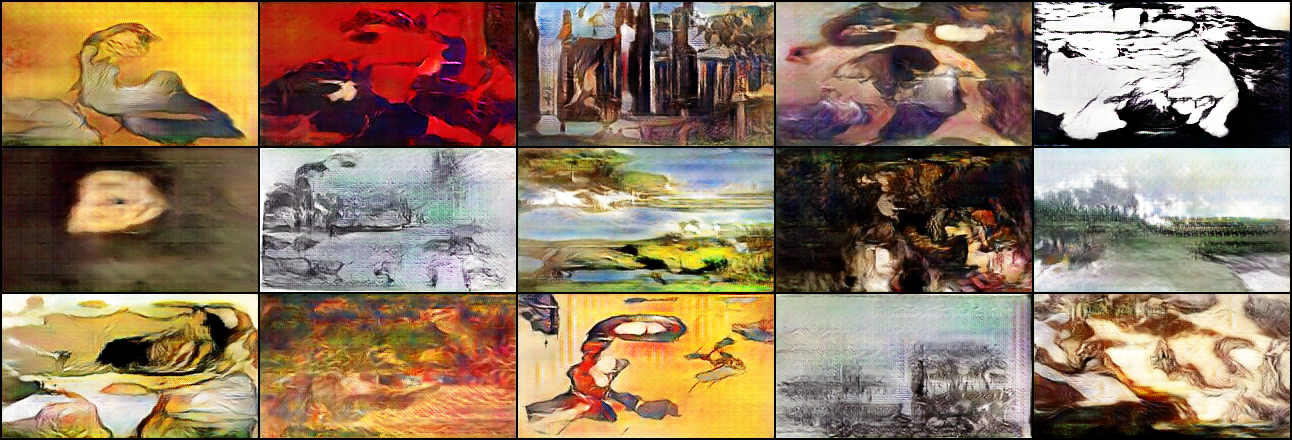}} 
    \subfigure[]{\includegraphics[width=0.99\textwidth]{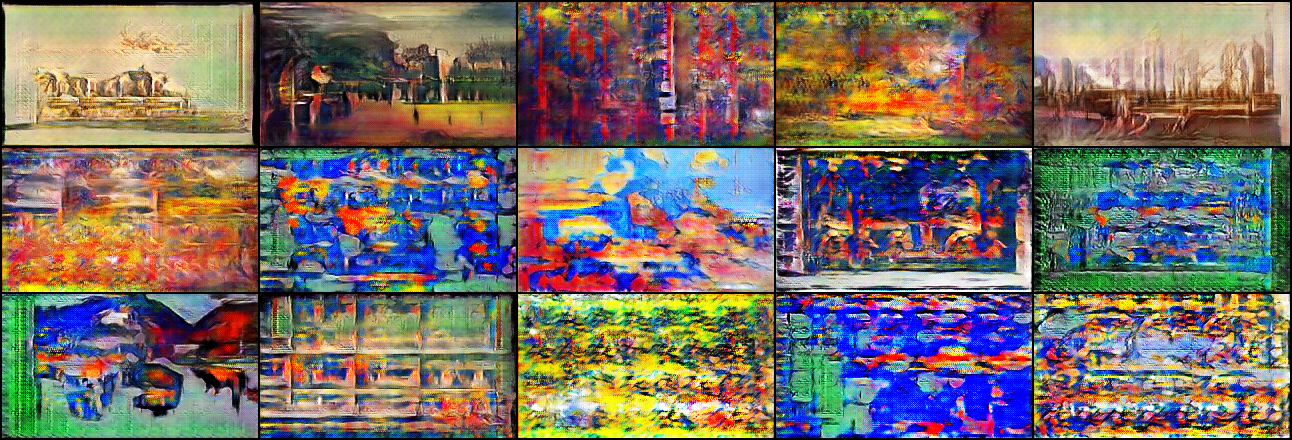}}
    \subfigure[]{\includegraphics[width=0.99\textwidth]{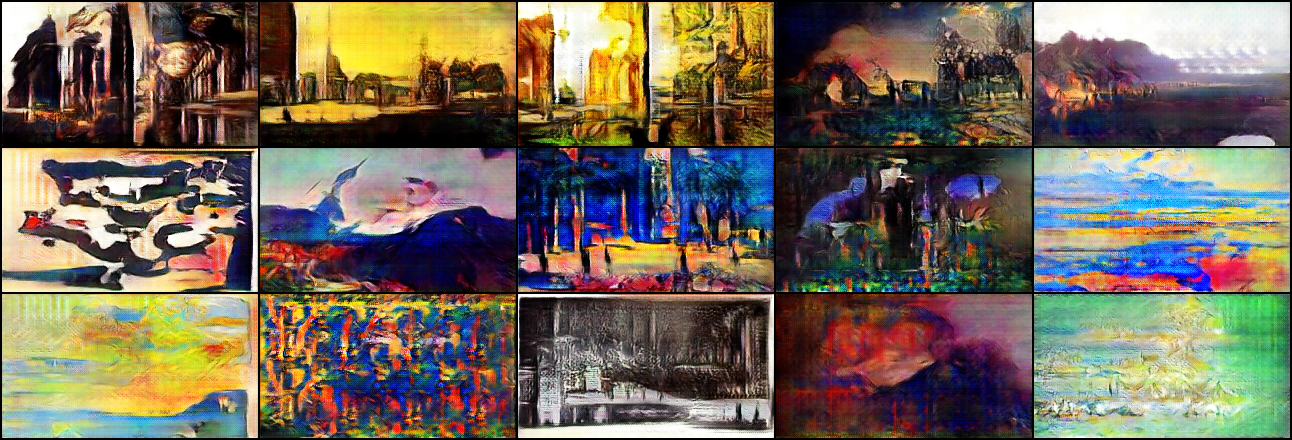}}
    \caption{(a) Population of random start images in the 1st generation, which served as starting point for both the automatic and the collaborative interactive evolution. (b) Population after 25 generations, evolved by the automatic aesthetic evaluation metric. (c) Population after 25 generations, evolved by the collaborative interactive evaluation. Visualisations of the entire evolutions can be found online. \\
    $\rightarrow$~Automatic evolution: \textcolor{blue}{\url{https://youtu.be/JCRx3Ih_0hA}} \\
    $\rightarrow$~Collaborative evolution: \textcolor{blue}{\url{https://youtu.be/rG_pLiX_UFo}}}
    \label{fig:Evolution}
\end{figure}

\newpage

\begin{figure}[h]
    \centering
    \subfigure[]{\includegraphics[width=0.99\textwidth]{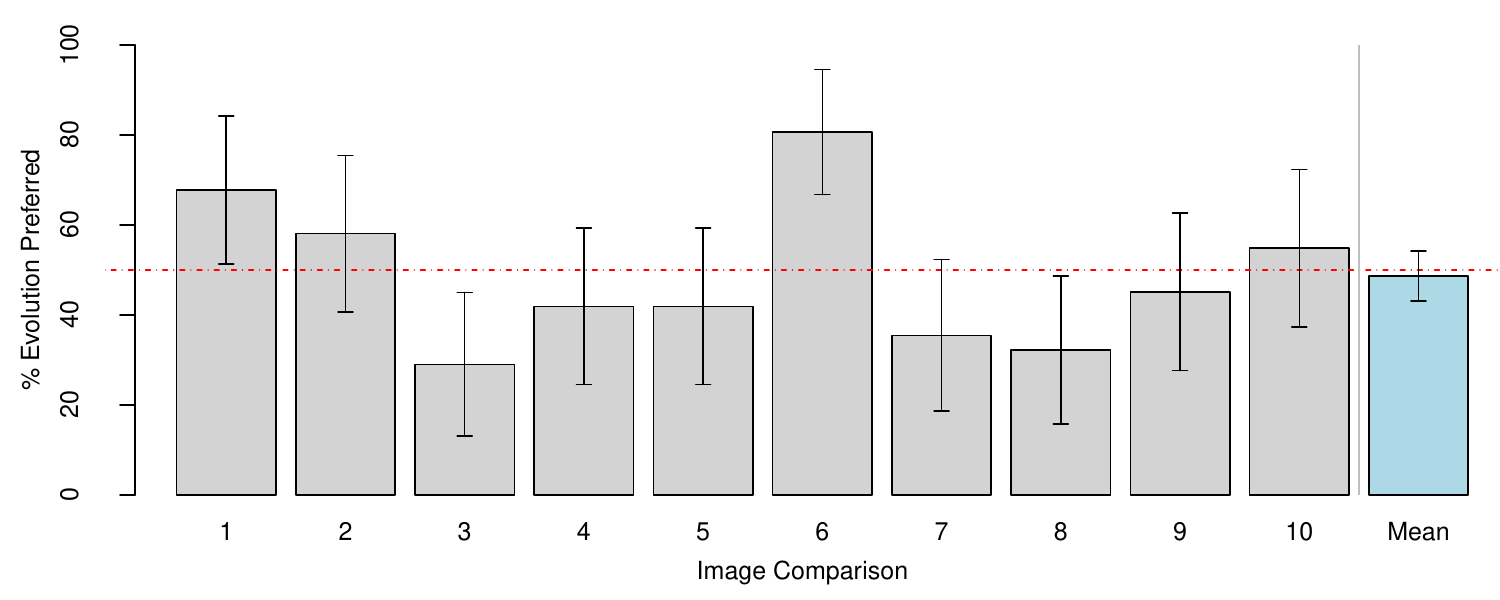}} 
    \subfigure[]{\includegraphics[width=0.99\textwidth]{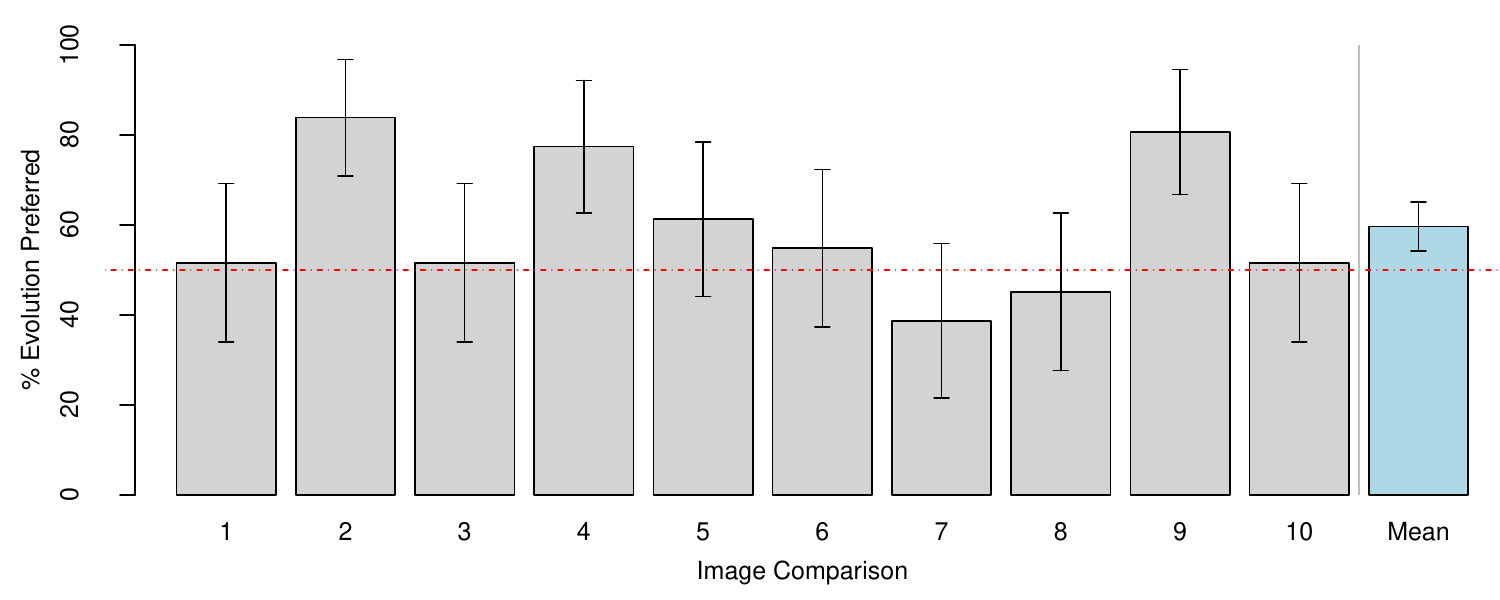}}
    \caption{(a) Proportion of participants preferring the results of the automatic aesthetic evolution over random images. (b) Proportion of participants preferring the results of the collaborative interactive evolution over random images. Both for all 10 comparisons as well as averaged over all images. The bars represent the standard errors.}
    \label{fig:BigEvaluation}
\end{figure}

\end{document}